\documentclass[runningheads]{llncs}

\usepackage{accv}

\usepackage{tabularx}
\usepackage[subtle]{savetrees}

\makeatletter
\let\LLNCSsubparagraph\subparagraph
\renewcommand\subparagraph{%
    \@startsection{subparagraph}{5}{\z@}%
    {-3pt}{-1em}%
    {\normalfont\normalsize\itshape}}
\makeatother

\usepackage{titlesec}

\makeatletter
\let\subparagraph\LLNCSsubparagraph
\makeatother

\titlespacing*{\section}{0pt}{8pt}{4pt}
\titlespacing*{\subsection}{0pt}{6pt}{3pt}
\titlespacing*{\subsubsection}{0pt}{5pt}{2pt}
\titlespacing*{\paragraph}{0pt}{4pt}{0.6em}

\AtBeginDocument{%
    \setlength{\abovedisplayskip}{8pt plus 2pt minus 2pt}%
    \setlength{\belowdisplayskip}{8pt plus 2pt minus 2pt}%
    \setlength{\abovedisplayshortskip}{6pt plus 2pt minus 2pt}%
    \setlength{\belowdisplayshortskip}{6pt plus 2pt minus 2pt}%
}
\usepackage{capt-of}
\usepackage{accvabbrv}

\usepackage{graphicx}
\usepackage{booktabs}

\usepackage[accsupp]{axessibility}  

\usepackage{hyperref}

\usepackage{orcidlink}

\begin{document}

\title{SCOPE: Semantic COdec Pre-Editing for Object-Preserving Compression} 

\titlerunning{SCOPE: Semantic COdec Pre-Editing}

\author{Silvia Dani\inst{1}\orcidlink{0009-0002-1656-3628} \and
Leonardo Galteri\inst{2}\orcidlink{0000-0002-7247-9407} \and
Marco Bertini\inst{1, 2}\orcidlink{0000-0002-1364-218X}}

\authorrunning{S.~Dani et al.}

\institute{University of Florence, Florence, Italy\\
\email{\{name.surname\}@unifi.it}\\
 \and
Small Pixels, Florence, Italy\\
\email{\{name.surname\}@small-pixels.com}}

\maketitle

\begin{abstract}
Image compression is increasingly used before machine vision systems, yet standard codecs are mainly optimized for human perception rather than downstream task accuracy. We propose SCOPE (Semantic COdec Pre-Editing), a conditional image pre-editor that modifies images before compression to improve the bitrate and detection accuracy trade-off. The model is conditioned on the codec family and compression severity, allowing a single trained network to adapt to multiple compression settings without retraining or loading codec-specific weights. Instead of uniformly preserving the whole image, the pre-editor learns to protect task-relevant regions while simplifying less important background content to improve compressibility. The resulting rate-accuracy curves show consistent gains across different codec families and detector architectures, together with strong transferability to unseen learned and wavelet-based compression families.


\end{abstract}

\section{Introduction}
\label{sec:introduction}

Image compression is a fundamental component of visual data transmission and storage. Most codecs, however, are designed according to human perceptual criteria, with the goal of preserving visual quality at low bitrate. This assumption is increasingly mismatched with machine-to-machine scenarios, where images are acquired, compressed, transmitted, and then consumed directly by computer vision systems rather than by human viewers. In these settings, the objective is not only visual fidelity, but the trade-off between bitrate and downstream task accuracy. Strong compression can remove fine object boundaries, alter textures, or introduce artifacts that degrade object detection while still spending bits on background content that is less relevant to the task. This reflects the known mismatch between perceptual distortion and machine-task performance~\cite{liu2019classification}.

This gap has motivated a growing body of work on image and video coding for machines. One line of research modifies or optimizes codecs with task-aware objectives, either by adapting standard codecs through task-driven bit allocation and rate-distortion optimization, or by training learned codecs with machine vision losses~\cite{choi2020task,chamain2021end,le2021image}. Another line of work transmits intermediate neural features rather than reconstructed images, with the aim of reducing bitrate while preserving the information needed by the downstream network~\cite{chen2019lossy,chen2019toward,singh2020end}. Although these approaches can improve rate-accuracy performance, they often require changes to the codec, to the transmitted representation, or to the downstream task pipeline. This limits their compatibility with standard image-compression systems and existing deployment settings.

A practical alternative is to insert a learned pre-processor before compression. Deep pre-editing has shown that images can be modified before encoding to improve compression efficiency while remaining compatible with standard codecs~\cite{talebi2021better}. For machine consumption, pre-processing methods edit the input image to suppress task-irrelevant information and preserve task-relevant content before compression~\cite{lu2024preprocessing,yang2023semantic,yang2024task}. This family of methods is attractive because the transmitted signal remains a standard compressed image and the approach can be combined with off-the-shelf codecs. However, existing pre-processing methods are often tied to a fixed codec, a fixed quality setting, or a task-specific configuration. Even task-switchable approaches focus on switching between downstream tasks, rather than adapting a single model across different codec families and compression severities. As a result, moving to a new codec or operating point may require retraining or relying on codec-specific pre-processors.

In this work, we address this limitation with SCOPE (Semantic COdec Pre-Editing), a conditional image pre-editor for compression-aware object detection. Instead of learning a separate pre-processor for each codec or bitrate regime, our model is conditioned on both the codec family and a normalized compression severity. This allows a single trained network to adapt its edits to multiple compression settings without retraining or loading codec-specific weights. The model predicts an attention map together with object and background residuals. Object regions are preserved through bounded residual edits, while background regions are simplified through a blurred-background branch to improve compressibility. This design reflects the asymmetry of object detection, where object-level cues should be protected more strongly than background texture.

The proposed formulation focuses on codec-aware pre-editing rather than task switching. The same model learns how to modify images for different compression families and severity levels, while maintaining a fixed downstream task. In this way, the conditioning mechanism captures compression-dependent editing behavior and enables transfer to unseen codec families.

The main contributions of this paper are as follows:
\begin{itemize}
    \item We propose a conditional pre-editing framework for object detection under image compression, where a single model adapts to different codec families and compression severities through learned conditioning.
    \item We introduce an attention-based object and background editing design that preserves object-relevant regions while simplifying less important background content to improve compressibility.
    \item We train the pre-editor with differentiable codec proxies and frozen detector supervision, enabling optimization of the rate-accuracy trade-off without modifying the downstream detector.
    \item We show that the learned model improves rate-accuracy performance across multiple codecs and detector architectures, and transfers to unseen compression families, including learned and wavelet-based codecs.
\end{itemize}
\noindent Code, configurations, and model weights are publicly available.\footnote{\texttt{\url{https://github.com/miccunifi/SCOPE}}}
\section{Related Work}
The classification-distortion-perception analysis of \cite{liu2019classification} has shown that classification accuracy, signal distortion and perceptual quality form a three-way trade-off that cannot be jointly optimized, meaning that improving perceptual realism can actively hurt a downstream classifier at fixed rate. That result reframes machine-oriented compression as a distinct operating regime rather than a tuning of the human one, and it is the premise that motivates our work. A more codec-specific formalization arrived with the rate-distortion theory for coding for machines of \cite{harell2022rate, harell2025rate}, which recasts the distortion term as a task-relevant fidelity measure and derives how bits should be allocated when the receiver is an analysis network. Together, these works establish that the object of optimization is a rate-accuracy curve, not a rate-PSNR curve, which is precisely the axis on which pre-editing methods, including the present paper, report their gains.

\subsubsection{Coding for machines}
The paradigm of video coding for machines was named and structured by Duan \etal~\cite{duan2020video}, who framed compression and analytics as a collaborative problem and set the agenda for jointly optimizing a codec against a downstream task. This development was soon followed by a series of fully end-to-end learned codecs. Le \etal~\cite{le2021image} introduced an image compression model trained directly under a downstream detection and recognition loss, aligning the compressed representation with vision-task performance. Chamain \etal~\cite{chamain2021end} conducted a comprehensive investigation of end-to-end optimized image compression tailored for machine consumption, analyzing how compression affects various automated vision tasks. The content-adaptive extension of \cite{le2021learned} dynamically adjusts the bit allocation according to the spatial and semantic characteristics of the image, explicitly optimizing this allocation with respect to a machine-centric objective function. A recurring theme is how much task supervision to bake into the codec: \cite{feng2022image} proposed ``omnipotent feature learning'' so that a single learned representation serves many tasks rather than overfitting the codec to one; the compact-visual-representation framework of \cite{yang2024video} proposes a unified representation compressed once and reused across analytics. A common limitation of these approaches is deployment cost: they all replace the standard codec with a trained network, so a new codec, bitrate regime, or task typically means retraining, in addition to the cost of redesigning existing video coding pipelines.

An alternative approach is to transmit neural features instead of pixels, trading codec compatibility for bitrate. The initial works in \cite{chen2019lossy, chen2019toward} propose to use intermediate-feature compression, while end-to-end compressible-features were proposed in \cite{singh2020end}: the goal is to learn representations that are cheap to code while remaining informative for the downstream network. Multi-scale feature compression with QP-adaptive channel selection has been proposed in \cite{yoon2023efficient}, scalable multilayer feature compression in \cite{chen2024end} and multiscale-importance-based bit allocation in \cite{liu2025multiscale}. More recent research has begun to abstract the notion of the transmitted representation itself, extending it across heterogeneous tasks and sensor modalities. Examples include the transformable feature framework TransVFC \cite{sun2026transvfc}, which seeks a unified task-agnostic feature space, and stereo feature coding for machines \cite{jin2026stereo}, which focuses on machine-centric representations of binocular data. However, this line of work implies that once the transmitted signal ceases to be a standard decodable image, it can no longer be rendered by standard image viewers or processed by existing image-based pipelines and tools. This incompatibility with the existing image ecosystems motivates the alternative design choice, that we have followed, of maintaining the output as a standards-based compressed image.
\smallskip
Because purely machine-oriented bitstreams sacrifice human viewability, a substantial effort targets bitstreams that serve both. Scalable designs place a machine-analysis base layer under a human-viewing enhancement layer: \cite{choi2022scalable} for video and the scalable image variant of \cite{wu2023end}. This line has matured into a standardization activity, with the open evaluation dataset of \cite{gao2022open} and the large-scale VCM benchmarking of \cite{marie2022video} supporting the MPEG effort, and recent assessments of the emerging MPEG VCM standard on segmentation \cite{phamdinh2025evaluating} and its bitstream/streaming structure \cite{park2025design}. A pragmatic sub-thread keeps the standard codec and adds neural pre/post modules around it like the learned intra-coding partitioning for VCM \cite{iino2024improving} and the lightweight post-processing ``task enhancement tiles'' \cite{partanen2025task}. These standard-compatible wrappers are the closest neighbors to pre-editing in spirit, but most operate inside or after the codec rather than as a pure, codec-agnostic front-end.

The idea that objects deserve more bits than background, central to the paper's attention-and-residual design, has a long codec-side history. Saliency- and object-driven bit allocation appears in the saliency-driven hierarchical learned codec of \cite{fischer2023saliency}, the saliency-map-guided codec of \cite{acmmm-2020, peng2024saliency}, the efficiency-oriented saliency approach of \cite{zhang2023efficient}, ROI-enabled learned coding \cite{ahonen2023region}, and object-analysis-driven use of a standard codec \cite{rek2023video}. Semantically disentangled coding ``telling the codec what is worth compressing'' \cite{liu2024tell} and ROI-packing for machine vision \cite{eimon2025roi} continue the theme, and recent CLIP-conditioned variants \cite{zhang2025blur,wang2025cgicm} use language-model semantics to decide what to preserve. The distinction from the present work is architectural: these methods realize the object/background asymmetry through codec-internal bit allocation, whereas the pre-editor realizes it in pixel space before an unmodified codec, via a predicted attention map that blends an object-preserving branch with a blurred background.
\vspace{-2pt}
\subsubsection{Preprocessing before standard codecs}

The closest family of approaches learns a preprocessing transform before a standard encoder, so the transmitted signal remains a conventional compressed image. Deep pre-editing~\cite{talebi2021better} uses this insertion point to improve the human-oriented rate-distortion trade-off. Earlier recognition-aware approaches also adapt the input for machine analysis: Choi \etal~\cite{choi2018high} study efficient compression for object detection, while Suzuki \etal~\cite{suzuki2019image} propose image pre-transformation for recognition-aware compression. SCOPE follows the same standard-compatible philosophy, but learns the transformation specifically for the downstream rate-accuracy trade-off.

More recent machine-oriented preprocessors suppress task-irrelevant content before encoding~\cite{lu2024preprocessing}, exploit task semantics~\cite{yang2023semantic}, or switch a shared preprocessor between downstream tasks~\cite{yang2024task}. Complementary work includes learned enhancement~\cite{ahonen2021learned}, video preprocessing under compression~\cite{zhao2024preprocessing}, GAN-based preprocessing~\cite{wu2024generative,wang2025regularized}, and diffusion-based recovery~\cite{dao2026gradient}. These methods motivate input adaptation for machine consumption, but do not explicitly condition one pre-editor on both codec family and compression severity. SCOPE instead targets codec switchability: a single set of weights is used across the evaluated compression families and operating points.
\vspace{-2pt}
\subsubsection{Learned compression backbones and conditioning}
The paper's own novelty, a single network that adapts across codec families and severities, is a conditioning problem, and what makes end-to-end training of any pre-codec module possible is a differentiable stand-in for the non-differentiable encoder. This paper builds on the differentiable-JPEG formulation whose subtleties were analyzed by Reich \etal~\cite{reich2024differentiable}, and on learned-codec proxies descended from the variational hyperprior of \cite{balle2018variationalimagecompressionscale} and the earlier nonlinear-transform-coding formulation of \cite{ball2016end}. This is the mechanism that lets detector gradients flow back through the compression stage to the pre-editor.

\section{Proposed Method}
\label{sec:method}

We propose SCOPE, a conditional pre-editing framework that modifies an input image before compression in order to preserve task-relevant content while making non-critical regions easier to compress. Given an RGB image $I \in [0,1]^{3 \times H \times W}$, a codec family identifier $f$, and a normalized compression severity $s \in [0,1]$, the model produces an edited image $I_{\mathrm{edit}}$. The edited image is then compressed by the target codec, or by a differentiable codec proxy during training. In our training setting, the codec family condition is sampled from Diff-JPEG~\cite{reich2024differentiable}, BMSHJ learned image compression~\cite{balle2018variationalimagecompressionscale}, and a BPG-like proxy. At inference time, the method requires only the input image and the desired codec condition $(f,s)$.
\vspace{-2pt}
\subsection{Model Architecture}
\label{subsec:model_architecture}

\begin{figure}[!htbp]
    \vspace{-2pt}
    \centering
    \makebox[\linewidth][c]{%
        \includegraphics[width=1.15\linewidth]{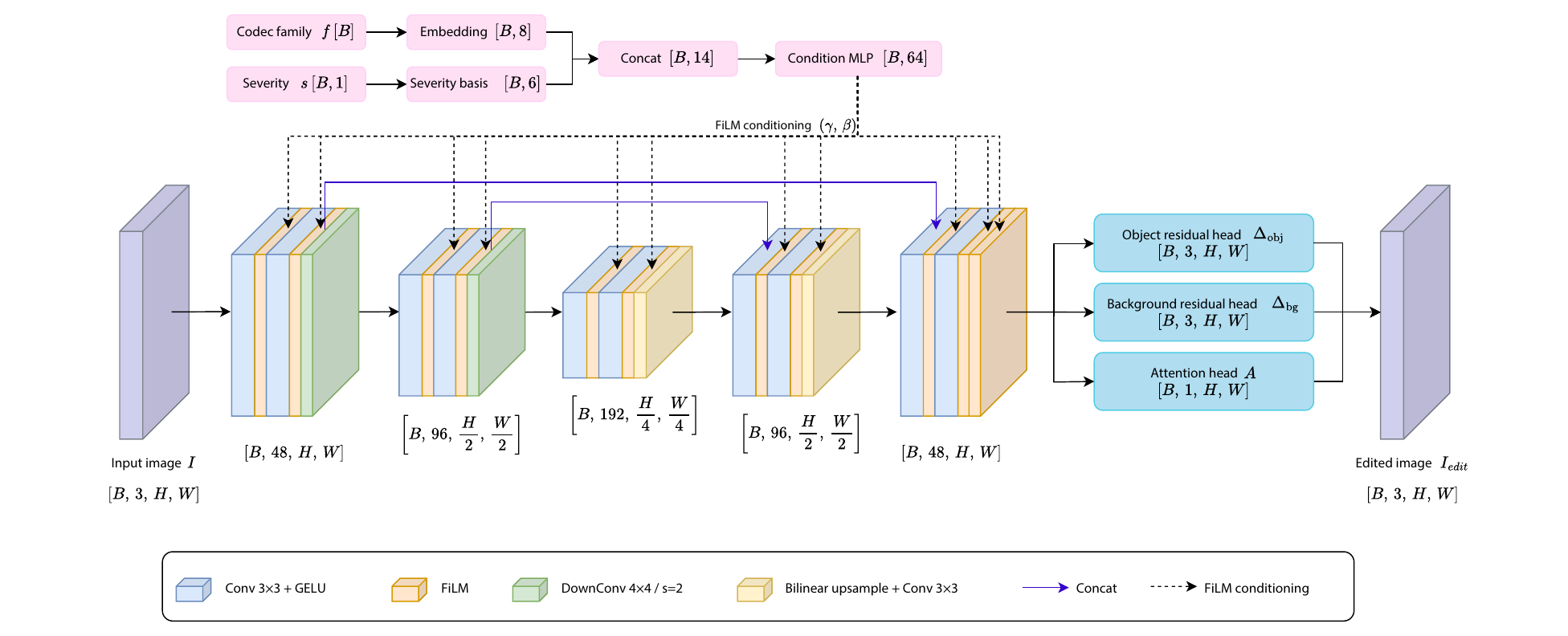}
    }
    \caption{
    Architecture of the proposed conditional pre-editor. The codec family $f$ and severity $s$ are encoded into a conditioning vector that modulates the U-Net-style image branch through FiLM layers. The network predicts an attention map $A$ and two residual maps, $\Delta_{\mathrm{obj}}$ and $\Delta_{\mathrm{bg}}$, which are used to construct the edited image $I_{\mathrm{edit}}$ by blending an object-preserving branch with a blurred-background branch.
    }
    \label{fig:model_pipeline}
    \vspace{-8pt}
\end{figure}

The architecture of the proposed conditional pre-editor is illustrated in Fig.~\ref{fig:model_pipeline}. The model consists of a conditioning branch and an image editing branch. The conditioning branch encodes the target codec family and compression severity into a compact conditioning vector. The codec family $f$ is represented by a learnable embedding $e_f \in \mathbb{R}^{8}$. The scalar severity $s$ is expanded to a fixed nonlinear severity basis:
\begin{equation}
    \phi(s)
    =
    \left[
    s,\,
    s^2,\,
    s^3,\,
    h_{0.50}(s),\,
    h_{0.75}(s),\,
    h_{0.90}(s)
    \right],
\end{equation}
where each high-severity basis term is defined as
\begin{equation}
    h_{\tau}(s)
    =
    \mathrm{clip}
    \left(
    \frac{s-\tau}{1-\tau},
    0,
    1
    \right).
\end{equation}
This expansion provides the condition MLP with both smooth polynomial severity cues and explicit high-severity indicators. This is useful because the effect of compression is not necessarily linear in the severity parameter: mild compression may require only small edits, while stronger compression can introduce rapidly increasing artifacts and may require a different pre-editing behavior. The high-severity terms allow the model to specialize its response when $s$ exceeds predefined thresholds, instead of forcing the MLP to infer these regimes from a single scalar input.

The family embedding and the severity basis are concatenated and processed by a condition MLP:
\begin{equation}
    c
    =
    \mathrm{MLP}
    \left(
    [e_f, \phi(s)]
    \right),
\end{equation}
producing a conditioning vector $c \in \mathbb{R}^{64}$.

The image branch follows a U-Net-style encoder-decoder architecture with skip concatenations. The encoder progressively reduces the spatial resolution while increasing the number of channels, producing multi-scale feature maps at resolutions $H \times W$, $H/2 \times W/2$, and $H/4 \times W/4$. The decoder restores the original resolution through bilinear upsampling followed by convolution, and concatenates the corresponding encoder features through skip connections. Each convolutional block is modulated by the conditioning vector through feature-wise linear modulation (FiLM)~\cite{perez2018film}:
\begin{equation}
    \mathrm{FiLM}(x,c) = x \odot (1+\gamma(c)) + \beta(c),
\end{equation}
where $\gamma(c)$ and $\beta(c)$ are predicted from the conditioning vector. This conditioning mechanism allows a single pre-editor to adapt its behavior to different codec families and compression severities.

Given the input image $I$, the pre-editor predicts an attention map $A$ and two residual maps, $\Delta_{\mathrm{obj}}$ and $\Delta_{\mathrm{bg}}$:
\begin{equation}
    A, \Delta_{\mathrm{obj}}, \Delta_{\mathrm{bg}}
    =
    P_{\theta}(I, f, s),
\end{equation}
where $A \in [0,1]^{1 \times H \times W}$ and
$\Delta_{\mathrm{obj}}, \Delta_{\mathrm{bg}} \in \mathbb{R}^{3 \times H \times W}$.
The object branch applies a bounded residual directly to the input image:
\begin{equation}
    I_{\mathrm{obj}}
    =
    \mathrm{clip}(I + \Delta_{\mathrm{obj}}, 0, 1).
\end{equation}
The background branch uses a blurred version of the input image as its base:
\begin{equation}
    I_{\mathrm{bg}}
    =
    \mathrm{clip}(\mathrm{Blur}(I) + \Delta_{\mathrm{bg}}, 0, 1).
\end{equation}
The final edited image is obtained by blending the two branches according to the predicted attention map:
\begin{equation}
    I_{\mathrm{edit}}
    =
    A \odot I_{\mathrm{obj}}
    +
    (1-A) \odot I_{\mathrm{bg}}.
    \label{eq:iedit}
\end{equation}
This design allows the model to preserve object-relevant regions through the object residual while simplifying background regions through the blurred-background branch.
\vspace{-2pt}
\subsection{Training Pipeline}
\label{subsec:training_pipeline}

\begin{figure}[t]
    \centering
    \includegraphics[width=\linewidth]{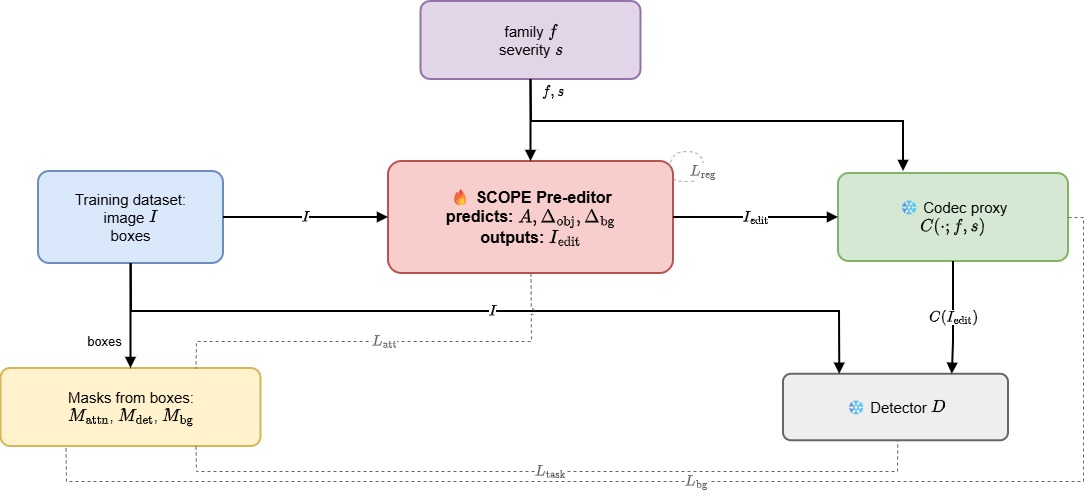}
    \caption{
    Training pipeline of the proposed method. The pre-editor is the only trainable module. The edited image is passed through a differentiable codec proxy, while ground-truth boxes are used only to construct training masks. A frozen detector provides a teacher-student consistency signal between the clean image and the compressed edited image. At inference time, masks and detector supervision are removed.
    }
    \label{fig:training_pipeline}
    \vspace{-8pt}
\end{figure}

The training pipeline is shown in Fig.~\ref{fig:training_pipeline}. During training, the edited image is passed through a differentiable codec proxy $C(\cdot;f,s)$, which approximates the effect of compression under codec family $f$ and severity $s$:
\begin{equation}
    \hat{I}
    =
    C(I_{\mathrm{edit}}; f, s),
\end{equation}
where $\hat{I}$ denotes the compressed edited image. Since the codec proxy is differentiable, gradients from the training objectives can flow through the compression stage and update the pre-editor. The codec proxy itself is kept fixed during pre-editor training.

\paragraph{Training setup.}
We train the pre-editor from scratch on COCO 2017~\cite{lin2014microsoft}. For each training sample, the codec family $f$ is sampled uniformly from
\begin{equation}
    \mathcal{F}
    =
    \{
    \texttt{Diff\_JPEG},
    \texttt{BMSHJ},
    \texttt{BPG\_like}
    \}.
\end{equation}

To enable end-to-end optimization, each codec family is represented during training by a differentiable proxy. For JPEG, we use a differentiable JPEG implementation that approximates transform coding and quantization while preserving gradient flow~\cite{reich2024differentiable,wallace1991jpeg}. For learned compression, we use the BMSHJ hyperprior model, which is already differentiable by construction and provides both reconstructed images and rate estimates~\cite{balle2018variationalimagecompressionscale}. Since no established differentiable proxy is commonly used for BPG, we implement a BPG-like surrogate that approximates its main compression effects through differentiable operations, allowing gradients to propagate through the compression stage during pre-editor training~\cite{bpg, sullivan2012overview}.

The normalized severity $s$ is sampled uniformly from a linear grid in $[0,1]$. We use no severity-dependent loss reweighting, so the codec family and severity determine the conditioning and proxy behavior but not the relative weights of the training losses.

Input images are cropped to $512 \times 512$, with object-centered crops sampled during training to expose the model to task-relevant regions. The pre-editor uses base channel width $48$, family embedding dimension $8$, conditioning dimension $64$, maximum object residual magnitude $0.09$, maximum background residual magnitude $0.03$, and a blurred-background replacement branch. We train for $40$ epochs with batch size $8$, learning rate $10^{-4}$, cosine scheduling, mixed-precision training, and gradient clipping.
\vspace{-2pt}
\paragraph{Training masks.}
Ground-truth bounding boxes are used to construct spatial masks that indicate which regions should be preserved and which regions can be simplified. We define three masks:
\begin{equation}
    M_{\mathrm{attn}}, \quad
    M_{\mathrm{det}}, \quad
    M_{\mathrm{bg}}.
\end{equation}
The attention mask $M_{\mathrm{attn}}$ is a tight object-region mask that is used to supervise the predicted attention map $A$. The detector mask $M_{\mathrm{det}}$ covers the object and nearby context regions and is used to restrict the detector-consistency loss to task-relevant areas. The background mask is defined as
\begin{equation}
    M_{\mathrm{bg}} = 1 - M_{\mathrm{det}},
\end{equation}
and identifies the regions where the model is encouraged to reduce visual complexity and improve compressibility.
\vspace{-2pt}
\paragraph{Detector supervision.}
To preserve task-relevant features after compression, we use a frozen detector $D$ as a teacher-student supervision signal. In this training configuration, $D$ is a Faster R-CNN detector~\cite{ren2015faster} with a ResNeXt-101-32x8d-FPN backbone~\cite{xie2017aggregated}, implemented in Detectron2~\cite{wu2019detectron2}. The clean image $I$ is passed through the detector to obtain teacher features, while the compressed edited image $\hat{I}$ is passed through the same frozen detector to obtain student features:
\begin{equation}
    F_T^\ell = D^\ell(I),
    \qquad
    F_S^\ell = D^\ell(\hat{I}),
\end{equation}
where $D^\ell(\cdot)$ denotes the feature map of the detector layer $\ell$. We extract features from the FPN levels and keep the detector frozen, so detector-based supervision updates only the pre-editor.
\vspace{-2pt}
\subsection{Training Objective}
\label{subsec:training_objective}

The model is trained using a combination of background-simplification, rate-oriented, attention-supervision, detector-consistency, and residual-regularization objectives. For readability, we group the loss terms according to their role:
\begin{equation}
\begin{aligned}
    \mathcal{L}
    &=
    \mathcal{L}_{\mathrm{bg}}
    +
    \mathcal{L}_{\mathrm{att}}
    +
    \mathcal{L}_{\mathrm{task}}
    +
    \mathcal{L}_{\mathrm{reg}},
    \label{eq:total_loss_grouped}
\end{aligned}
\end{equation}
where
\begin{equation}
\begin{aligned}
    \mathcal{L}_{\mathrm{bg}}
    &=
    \lambda_{\mathrm{bg\_flat}}\mathcal{L}_{\mathrm{bg\_flat}}
    +
    \lambda_{\mathrm{bg\_smooth}}\mathcal{L}_{\mathrm{bg\_smooth}}
    +
    \lambda_{\mathrm{rate}}\mathcal{L}_{\mathrm{rate}},
    \\[2mm]
    \mathcal{L}_{\mathrm{att}}
    &=
    \lambda_{\mathrm{attn}}\mathcal{L}_{\mathrm{attn}}
    +
    \lambda_{\mathrm{attn\_sparse}}\mathcal{L}_{\mathrm{attn\_sparse}}
    +
    \lambda_{\mathrm{attn\_area}}\mathcal{L}_{\mathrm{attn\_area}},
    \\[2mm]
    \mathcal{L}_{\mathrm{task}}
    &=
    \lambda_{\mathrm{det}}\mathcal{L}_{\mathrm{det}},
    \\[2mm]
    \mathcal{L}_{\mathrm{reg}}
    &=
    \lambda_{\Delta}\mathcal{L}_{\Delta}
    +
    \lambda_{\Delta\mathrm{TV}}\mathcal{L}_{\Delta\mathrm{TV}}.
\end{aligned}
\label{eq:loss_groups}
\end{equation}
\paragraph{Background simplification and rate regularization.}
The background group $\mathcal{L}_{\mathrm{bg}}$ encourages non-object regions to be easier to compress while avoiding disruptive artifacts. We use a blur-based flatness term and a smoothness term:
\begin{equation}
\begin{aligned}
    \mathcal{L}_{\mathrm{bg\_flat}}
    &=
    \frac{
    \left\|
    M_{\mathrm{bg}} \odot
    \left(\hat{I} - \mathrm{Blur}(\hat{I})\right)
    \right\|_1
    }{
    \sum M_{\mathrm{bg}} + \epsilon
    },
    \qquad
    \mathcal{L}_{\mathrm{bg\_smooth}}
    =
    \frac{
    \left\|
    M_{\mathrm{bg}} \odot \nabla \hat{I}
    \right\|_1
    }{
    \sum M_{\mathrm{bg}} + \epsilon
    }.
\end{aligned}
\end{equation}
The rate loss $\mathcal{L}_{\mathrm{rate}}$ further encourages compressibility by penalizing structures that increase the expected coding cost. For learned compression, this can be estimated from the codec proxy rate term, while for hand-crafted proxies, we use differentiable smoothness and rate surrogates.
\vspace{-2pt}
\paragraph{Attention supervision.}
The attention group $\mathcal{L}_{\mathrm{att}}$ supervises the predicted attention map and regularizes its spatial support. The main attention loss is defined by
\begin{equation}
    \mathcal{L}_{\mathrm{attn}}
    =
    \mathcal{L}_{\mathrm{BCE}}(A, M_{\mathrm{attn}})
    +
    \mathcal{L}_{\mathrm{Dice}}(A, M_{\mathrm{attn}}).
\end{equation}
We additionally use sparsity and area regularization terms, $\mathcal{L}_{\mathrm{attn\_sparse}}$ and $\mathcal{L}_{\mathrm{attn\_area}}$, to discourage unnecessary high attention outside object regions.
\vspace{-2pt}
\paragraph{Detector consistency.}
The task group $\mathcal{L}_{\mathrm{task}}$ preserves detector-relevant features after pre-editing and compression. We compare teacher and student detector features inside the detector mask using a Smooth L1 feature loss:
\begin{equation}
    \mathcal{L}_{\mathrm{det}}
    =
    \sum_{\ell}
    \frac{
    \left\|
    M_{\mathrm{det}}^\ell \odot
    \rho
    \left(
    F_S^\ell - F_T^\ell
    \right)
    \right\|_1
    }{
    \sum M_{\mathrm{det}}^\ell + \epsilon
    },
\end{equation}
where $\rho(\cdot)$ denotes the Smooth L1 penalty and $M_{\mathrm{det}}^\ell$ is the detector mask resized to the spatial resolution of layer $\ell$.
\vspace{-2pt}
\paragraph{Residual regularization.}
The regularization group $\mathcal{L}_{\mathrm{reg}}$ keeps the predicted edits bounded and spatially smooth. We penalize the magnitude of the object and background residuals:
\begin{equation}
\begin{aligned}
    \mathcal{L}_{\Delta}
    &=
    \|\Delta_{\mathrm{obj}}\|_1
    +
    \|\Delta_{\mathrm{bg}}\|_1,
    \qquad
    \mathcal{L}_{\Delta\mathrm{TV}}
    =
    \mathrm{TV}(\Delta_{\mathrm{obj}})
    +
    \mathrm{TV}(\Delta_{\mathrm{bg}}).
\end{aligned}
\end{equation}
where $\mathrm{TV}(\cdot)$ denotes total variation regularization.
\vspace{-2pt}
\subsection{Inference}
\label{subsec:inference}

At inference time, the training-only supervision is removed. No bounding boxes, masks, or detector features are required. Given an image $I$, a codec family $f$, and a target severity $s$, the pre-editor produces $I_{\mathrm{edit}}$, which is then encoded by the target codec. Thus, the method remains image-only at deployment while having been trained to preserve detector-relevant content under compression.

Although our experiments focus on object detection, SCOPE is not architecturally tied to detection. The training formulation only requires a frozen downstream task model to provide task-aware supervision; in principle, the detector could be replaced by a segmentation or pose estimator without modifying the codec-facing pre-editor architecture. We therefore keep the empirical conclusions of this work specific to object detection while noting that the formulation itself is task-model agnostic.
\vspace{-2pt}
\section{Experiments}
\label{sec:experiments}

We evaluate the proposed pre-editor by measuring the detection mAP obtained after compression as a function of the resulting bitrate. For each codec, we compare the detector performance on compressed images without pre-editing against the performance obtained after applying the proposed pre-editor before compression. We report curves for three detector architectures: Faster R-CNN with a ResNeXt-101-32x8d-FPN backbone~\cite{ren2015faster,xie2017aggregated}, RT-DETR-l~\cite{zhao2024detrs}, and YOLO26x~\cite{yolo26}. This allows us to assess both rate-accuracy behavior and cross-detector generalization.

\paragraph{Evaluation setup.}
All quantitative experiments use the 5,000-image COCO 2017 validation split. For the codec families used during training, we evaluate JPEG at quality values $\{10,15,20,30,40,50\}$, BPG at QPs $\{28,31,34,37,41,44\}$, and BMSHJ at quality levels $\{1,2,3,4,5\}$. For zero-shot transfer, JPEG2000 uses OpenJPEG rate parameters $\{32,48,64,96,128,192\}$ and Cheng uses quality levels $\{1,\ldots,6\}$. The severity condition linearly normalizes the operating parameter within each codec-specific range, with the direction chosen so that larger $s$ always denotes stronger compression. Importantly, the bitrate reported in the rate-accuracy curves is measured from the actual encoded output rather than inferred from the nominal codec parameter. The supplementary reports $7.93$ ms for the network forward pass and $11.81$ ms total pre-editing time per image, corresponding to $84.7$ FPS including tensor preparation and conditioning.
\vspace{-2pt}
\subsection{Results}

\begin{figure}[!htbp]
    \centering
    \includegraphics[width=0.32\linewidth]{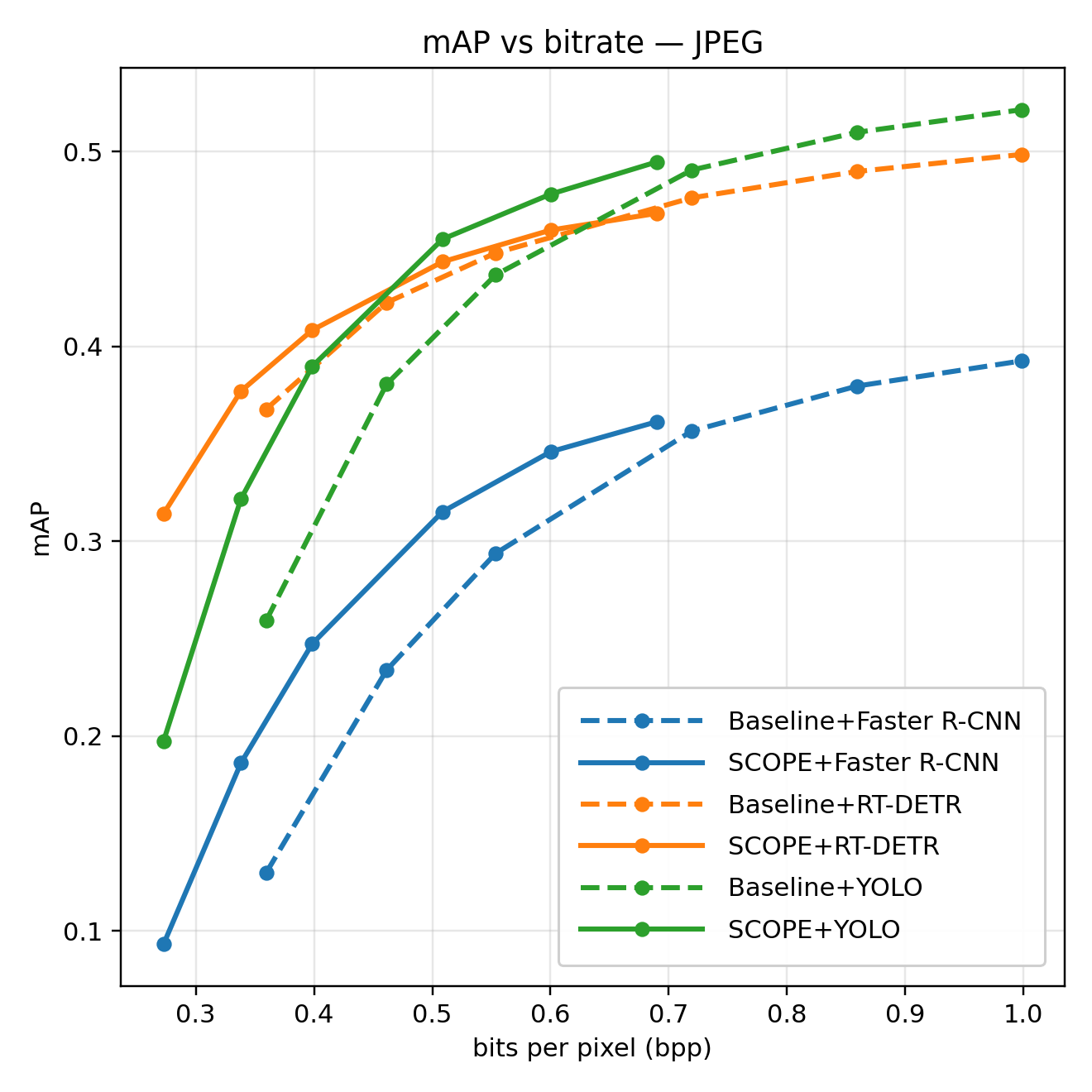}
    \includegraphics[width=0.32\linewidth]{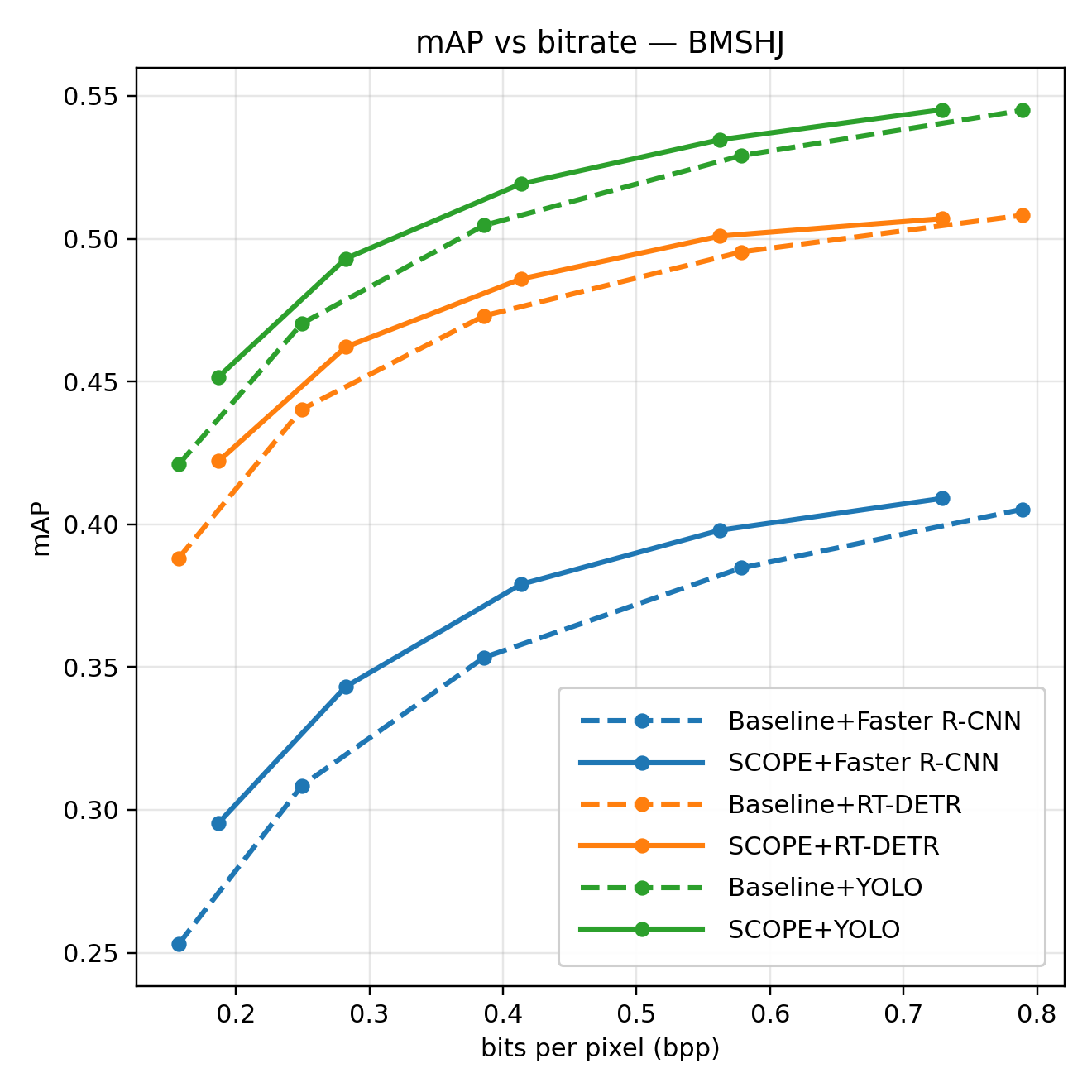}
    \includegraphics[width=0.32\linewidth]{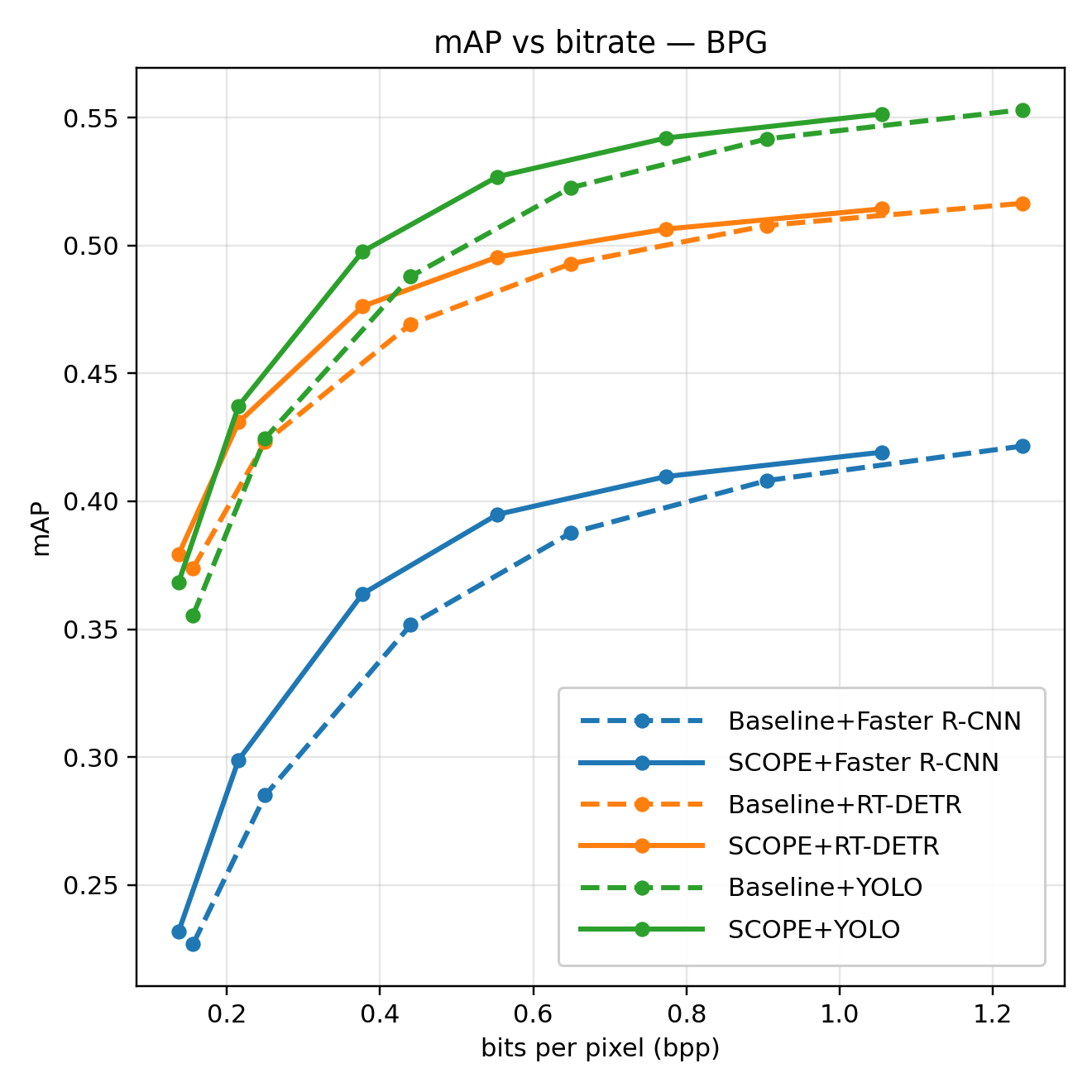}
    \caption{
    mAP versus bitrate for the codec families used during training: JPEG, BMSHJ, and BPG. Results are shown for Faster R-CNN, RT-DETR-l, and YOLO26x.
    }
    \label{fig:seen_codec_results}
    \vspace{-8pt}
\end{figure}

Fig.~\ref{fig:seen_codec_results} reports results for JPEG, BMSHJ, and BPG. Across the three codecs, SCOPE generally shifts the rate-accuracy curves toward higher mAP at comparable bitrate or comparable mAP at lower bitrate. For BMSHJ and BPG, the improvement is consistent across Faster R-CNN, RT-DETR-l, and YOLO26x, showing that the learned edits are not limited to the detector used for training.

JPEG is more detector-dependent. The gain is clear for Faster R-CNN and YOLO26x ($-15.88\%$ and $-14.92\%$ BDBR), whereas for RT-DETR-l it is concentrated in the low-bitrate regime and the curves converge as compression weakens ($-5.71\%$). This behavior is consistent with the role of pre-editing: when the baseline codec already preserves most task-relevant information, there is less room to improve the detector at a fixed rate.

Although training uses Faster R-CNN feature supervision, gains also transfer to RT-DETR-l and YOLO26x across BMSHJ, BPG, the unseen Cheng codec (Sec.~\ref{subsec:zero_shot_codec_transferability}), and part of the JPEG range. This cross-detector behavior suggests that the pixel-space edits preserve object-relevant evidence shared across two-stage, transformer-based, and one-stage detectors rather than overfitting to a single architecture.
\vspace{-2pt}

We summarize complete rate-accuracy curves with BDBR~\cite{bjontegaard2001calculation}, using AP as quality and bitrate as rate, over each pair's common AP interval without extrapolation. Negative values denote bitrate savings at equal accuracy. SCOPE averages $-15.98\%$ over the nine JPEG/BPG/BMSHJ-detector pairs, all of which are negative (Table~\ref{tab:conditioning_all}).

\subsection{Qualitative Comparison}
\label{subsec:qualitative_comparison}

\begin{figure}[t]
    \centering
    \includegraphics[width=\linewidth]{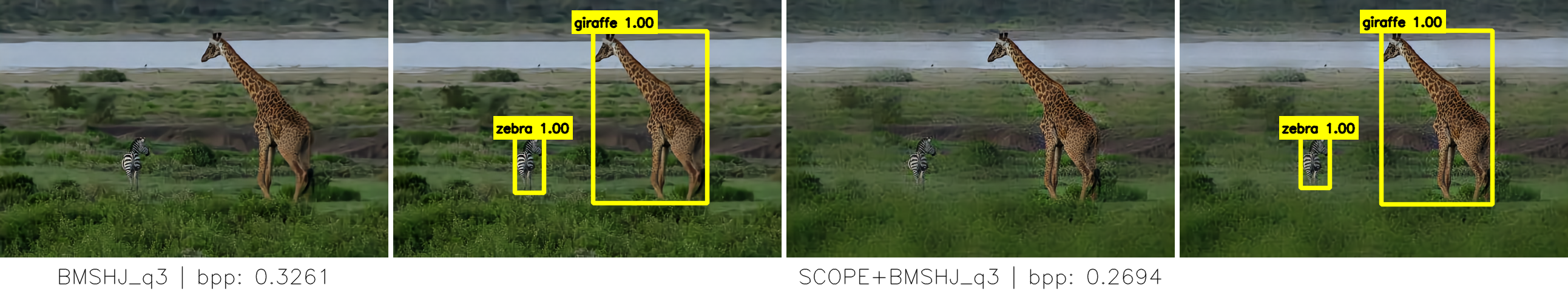}\\
    \includegraphics[width=\linewidth]{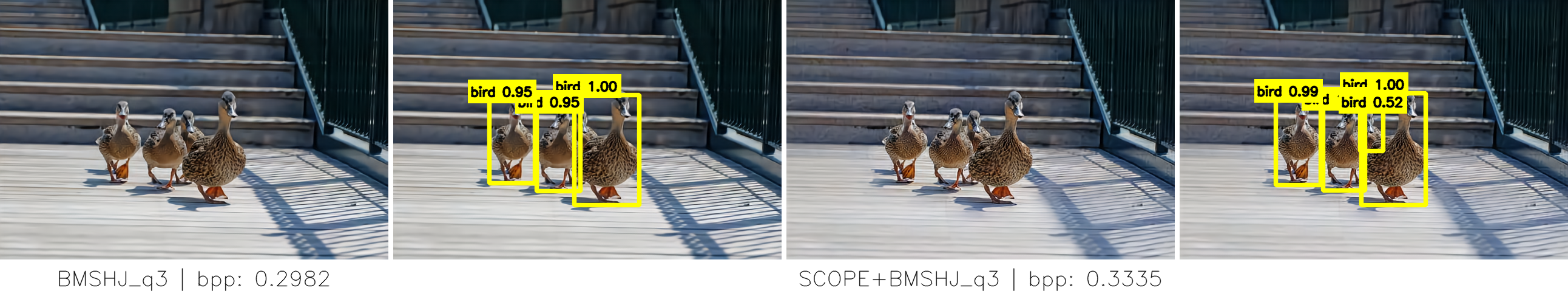}\\
    \includegraphics[width=\linewidth]{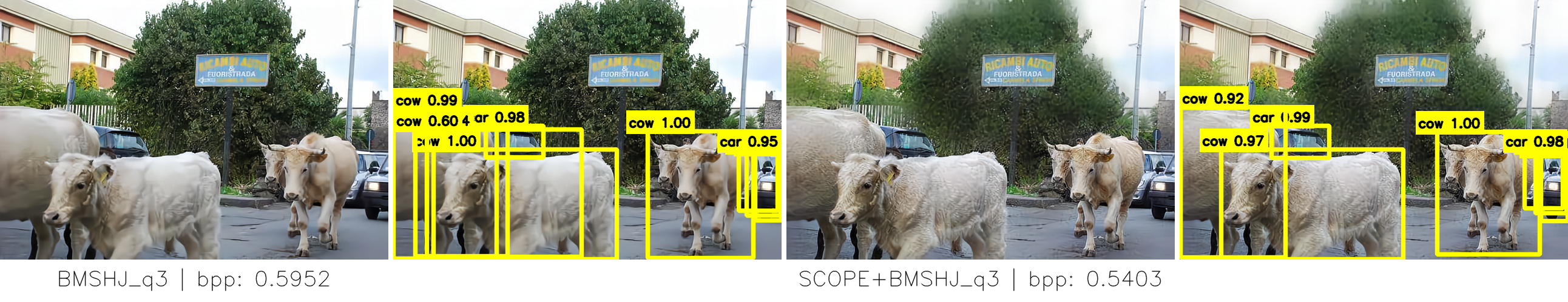}
    \caption{
    Qualitative comparison for BMSHJ compression at quality level q3. From left to right: baseline compressed image without detections and with Faster R-CNN detections, SCOPE pre-edited compressed image without detections, and with Faster R-CNN detections.
    }
    \label{fig:qualitative_BMSHJ}
    \vspace{-8pt}
\end{figure}

Fig.~\ref{fig:qualitative_BMSHJ} shows qualitative examples of BMSHJ compression. Across the examples, SCOPE does not enforce a single visual outcome but adapts the pre-editing to the detection-rate trade-off. In many cases, it reduces the bitrate while preserving the detections obtained after baseline compression, or even suppressing false positives. In other cases, it may slightly increase the bitrate, but recovers objects that are missed by the compressed baseline. Overall, these examples illustrate the intended behavior of the method: background content is modified to improve compressibility when possible, while object regions are kept sufficiently informative or selectively enhanced to support object detection.
\vspace{-2pt}

\subsection{Zero-Shot Codec Transferability}
\label{subsec:zero_shot_codec_transferability}

\begin{figure}[t]
    \centering
    \includegraphics[width=0.33\linewidth]{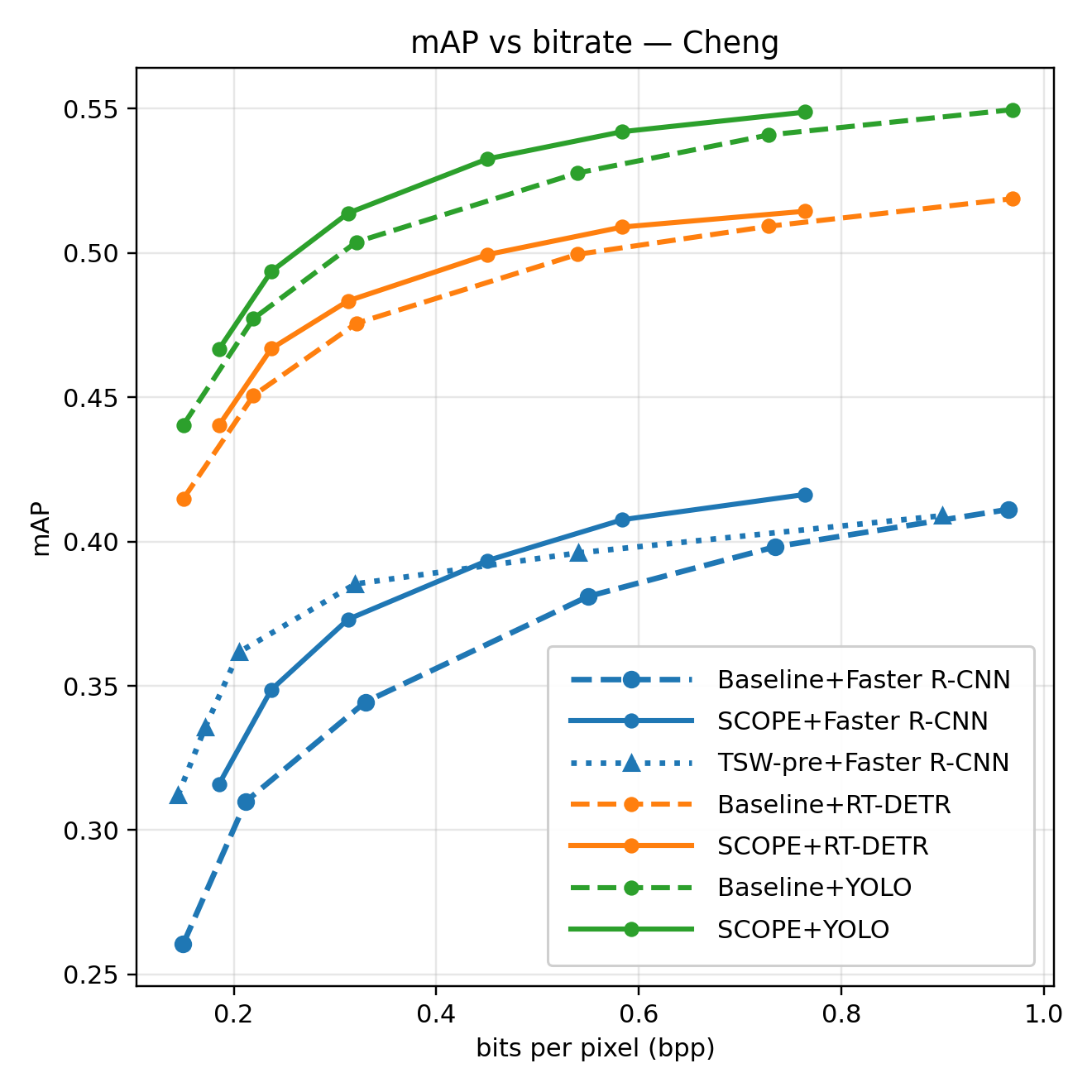}
    \includegraphics[width=0.33\linewidth]{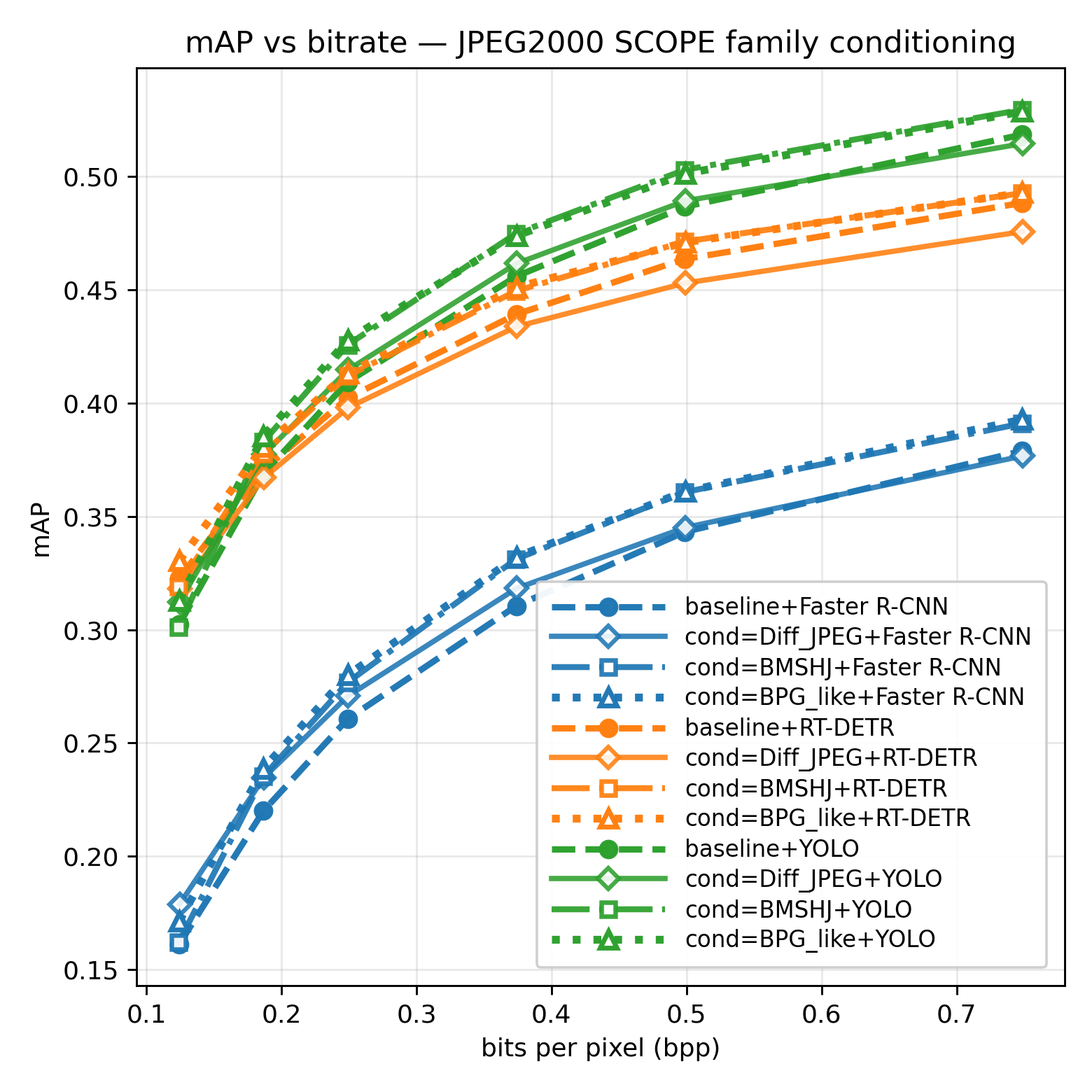}
    \caption{
    Zero-shot codec transfer. Left: mAP versus bitrate for the unseen Cheng
    learned codec. Right: JPEG2000, where the same pretrained SCOPE model is
    evaluated under each of the three available seen-family conditions.
    }
    \label{fig:unseen_codec_results}
    \vspace{-2pt}
\end{figure}

We evaluate transfer to two codecs that are not explicit training targets: Cheng~\cite{cheng2020learnedimagecompressiondiscretized}, an unseen learned codec, and JPEG2000~\cite{skodras2002jpeg}, whose wavelet-based family is absent from training (Fig.~\ref{fig:unseen_codec_results}). On Cheng, SCOPE improves direct compression for all three detectors. For Faster R-CNN it obtains $-20.66\%$ BDBR zero-shot, while the Cheng-trained task-switchable pre-processor of Yang \etal~\cite{yang2023semantic} (TSW-pre) obtains $-31.32\%$. Thus, the codec-specific method remains stronger on the codec seen during its training, whereas SCOPE retains substantial savings without Cheng-specific optimization. On the matched BPG/Faster R-CNN setting, SCOPE obtains $-21.74\%$ compared with $-19.50\%$ for Lu \etal~\cite{lu2024preprocessing}. The TSW-pre result is reconstructed from its published curve, and each BDBR is evaluated only over the corresponding valid common AP interval.

JPEG2000 tests a different form of transfer because its wavelet-based family is unseen. We do not use target-codec validation to choose a family condition. Instead, the same pretrained model is evaluated without retraining under all three available conditions: \texttt{Diff\_JPEG}, \texttt{BMSHJ}, and \texttt{BPG\_like}. Their mean BDBRs are $-1.35\%$, $-9.97\%$, and $-11.92\%$, respectively (Table~\ref{tab:jp2_conditioning}), with BPG-like strongest for all three detectors. The common ordering across detector architectures indicates that the conditioning effect is not specific to a single detector.

Differentiable codecs are used only as training proxies; all evaluation curves use the actual codecs. We assume no bitstream or numerical parameter equivalence between a proxy and a real codec. Severity represents codec-normalized relative compression strength, while the family condition represents the degradation type.

\vspace{-1pt}

\vspace{-1mm}
\noindent
\begin{table*}[t]
\centering

\begin{minipage}[!htbp]{0.56\textwidth}
\centering
\captionof{table}{JPEG2000 zero-shot BDBR (\%).}
\label{tab:jp2_conditioning}

\scriptsize
\setlength{\tabcolsep}{3pt}
\renewcommand{\arraystretch}{0.95}

\begin{tabular}{lrrrr}
\toprule
Condition &
\shortstack{Faster\\R-CNN} &
\shortstack{RT-\\DETR-l} &
YOLO26x &
Mean \\
\midrule

Diff-JPEG
& -6.82
& +7.00
& -4.24
& -1.35 \\

BMSHJ
& -11.60
& -7.42
& -10.89
& -9.97 \\

BPG-like
& \textbf{-13.73}
& \textbf{-9.74}
& \textbf{-12.28}
& \textbf{-11.92} \\

\bottomrule
\end{tabular}

\end{minipage}
\hfill
\begin{minipage}[!htbp]{0.40\textwidth}
\centering
\captionof{table}{Matched Faster R-CNN BDBR (\%).}
\label{tab:published_baselines}

\scriptsize
\setlength{\tabcolsep}{4pt}
\renewcommand{\arraystretch}{0.95}

\begin{tabular}{lllr}
\toprule
Codec & Method & Train & BDBR \\
\midrule

BPG
& Lu \etal~\cite{lu2024preprocessing}
& Seen
& -19.50 \\

BPG
& SCOPE
& Seen
& \textbf{-21.74} \\

\midrule

Cheng
& TSW-pre~\cite{yang2023semantic}
& Seen
& \textbf{-31.32} \\

Cheng
& SCOPE
& Unseen
& -20.66 \\

\bottomrule
\end{tabular}

\end{minipage}

\vspace{-1mm}
\end{table*}

\vspace{2mm}

\subsection{Ablation Study}
\label{subsec:ablation_study}

\begin{figure}[t]
    \centering
    \includegraphics[width=0.32\linewidth]{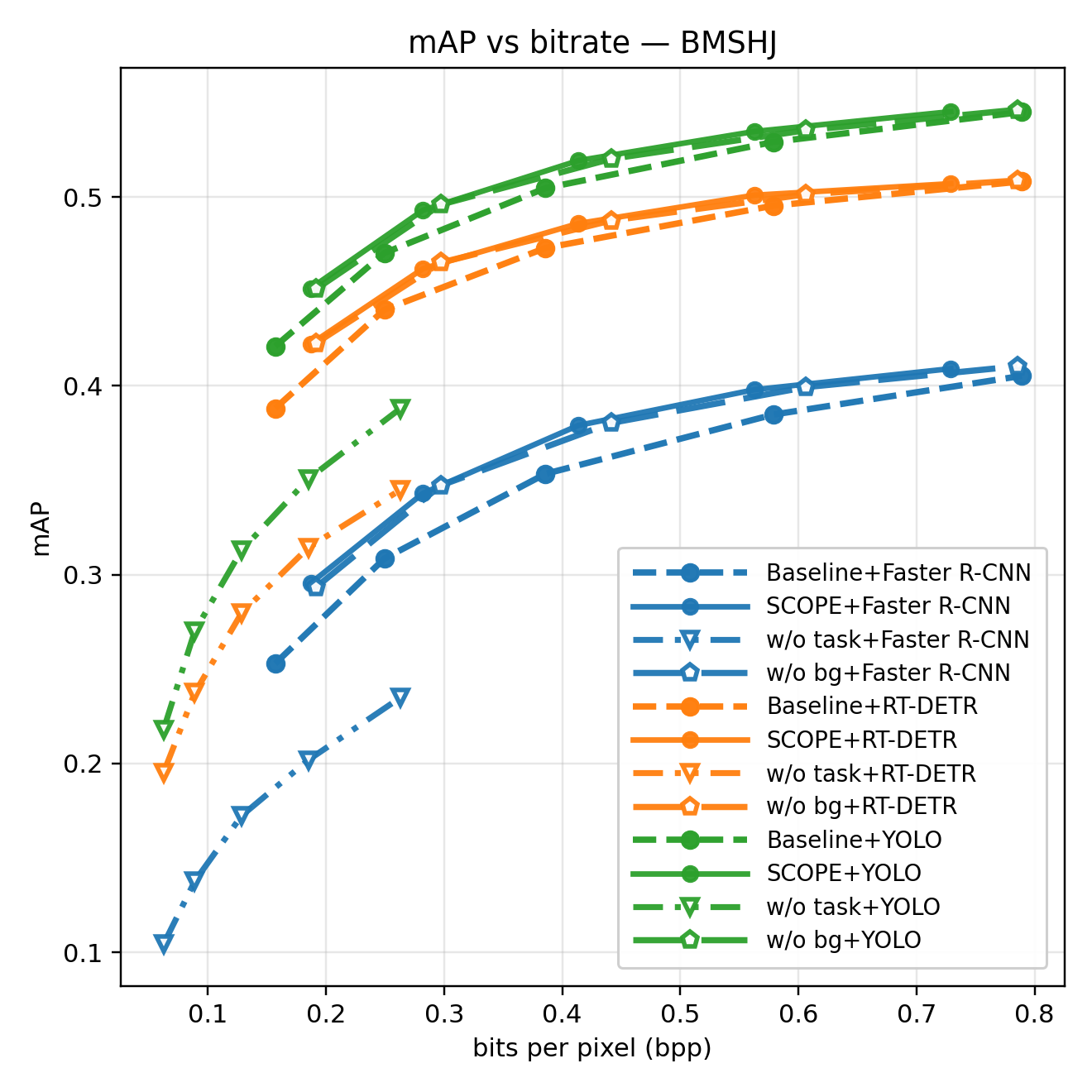}
    \includegraphics[width=0.32\linewidth]{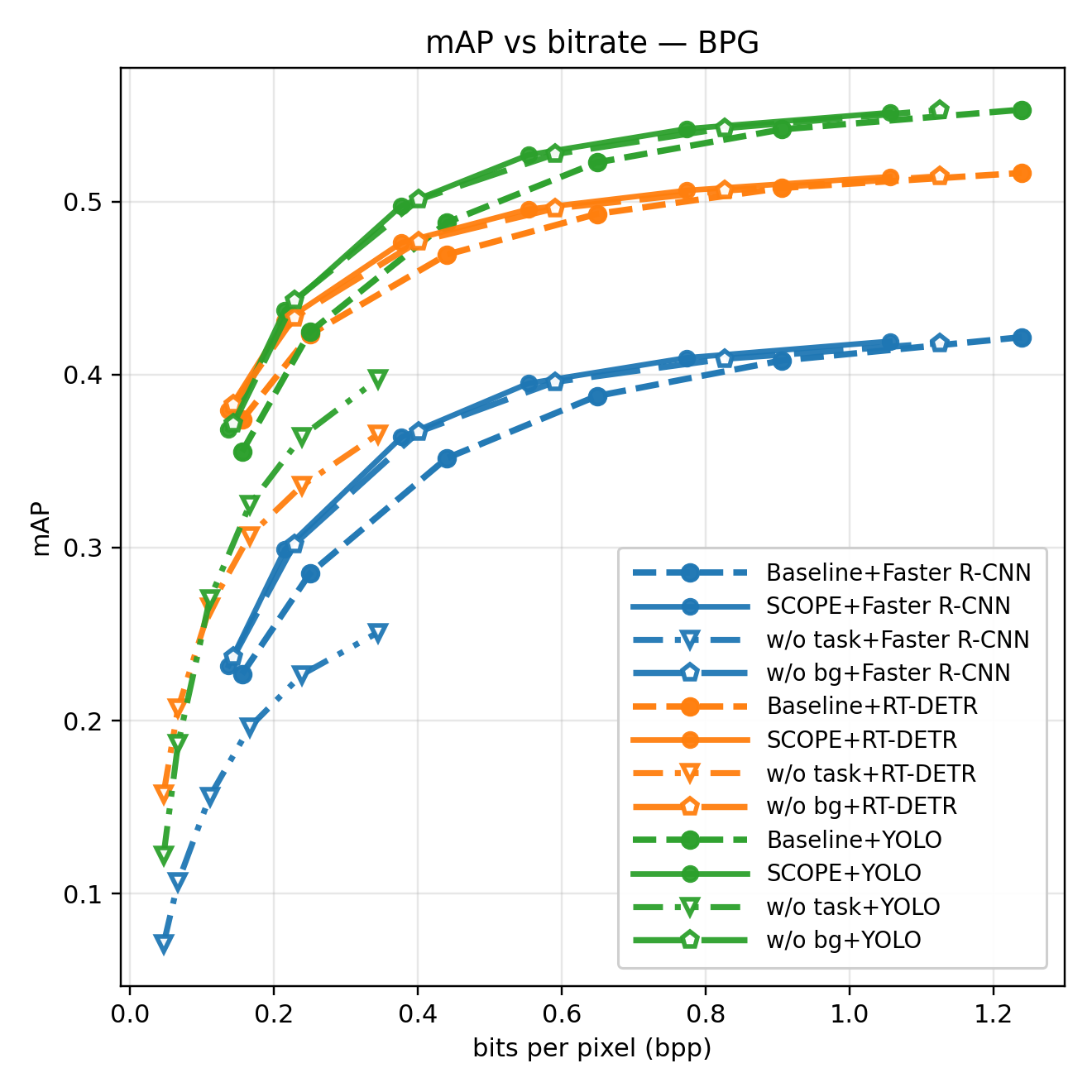}
    \includegraphics[width=0.32\linewidth]{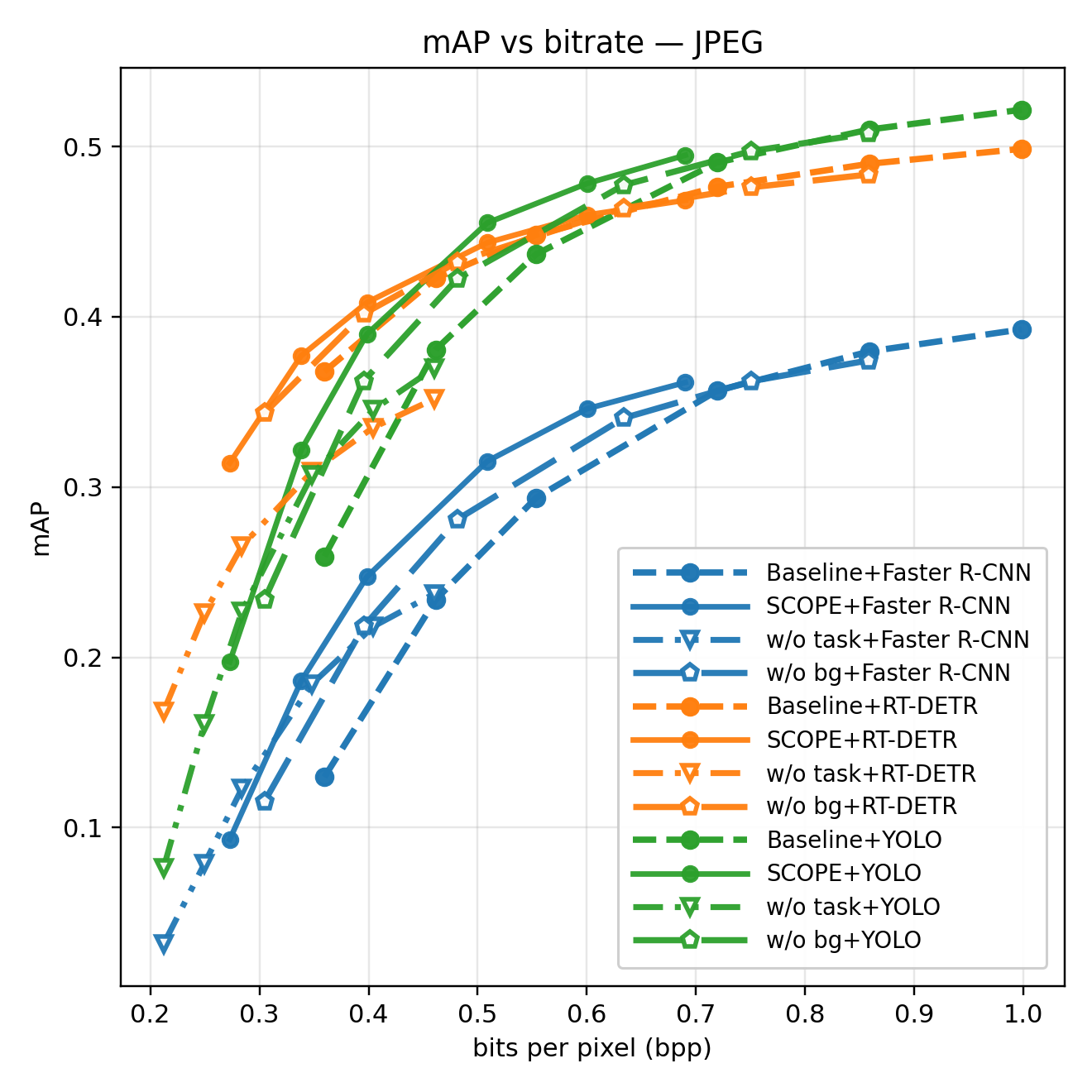}
    \caption{
    Ablation study on the training objectives. Dashed curves correspond to
    direct compression without pre-editing, solid circle curves to full SCOPE,
    downward-triangle curves to \emph{w/o task}, and pentagon curves to
    \emph{w/o bg/rate}. The task-aware loss is essential for preserving
    detection accuracy, while the background and rate group improves the
    compression-aware rate-accuracy trade-off.
    }
    \label{fig:ablation_study}
    \vspace{-2pt}
\end{figure}

\paragraph{Training objectives.}
Fig.~\ref{fig:ablation_study} compares full SCOPE with variants removing the task-aware loss $\mathcal{L}_{\mathrm{task}}$ (\emph{w/o task}) or the background simplification and rate group $\mathcal{L}_{\mathrm{bg}}$ (\emph{w/o bg/rate}). Removing $\mathcal{L}_{\mathrm{task}}$ has the strongest negative effect. Although the resulting curves can move toward lower bitrate, this comes with a substantial loss of detection accuracy and the variant often falls below direct compression. Bitrate reduction alone is therefore insufficient: detector-guided supervision is needed to prevent the pre-editor from discarding information that matters to the task.

The \emph{w/o bg/rate} variant remains substantially closer to full SCOPE, indicating that task supervision still protects object-relevant evidence. Its rate-accuracy behavior is nevertheless less stable across codecs and detectors. The two groups therefore play complementary roles: $\mathcal{L}_{\mathrm{task}}$ protects task evidence, while background simplification and rate regularization encourage non-object regions to consume fewer unnecessary bits.

\paragraph{Codec-family and severity conditioning.}
We compare full SCOPE with four alternatives (Table~\ref{tab:conditioning_all}): \emph{Family} removes severity conditioning, \emph{Severity} removes the family condition, \emph{Per-codec} trains a separate unconditioned specialist for each codec, and \emph{Per-codec+Sev.} adds severity conditioning to each codec-specific specialist. \emph{Base} reports BDBR relative to direct compression, while \emph{SCOPE} reports a direct pairwise comparison with the full model. These latter values are recomputed over the valid common AP interval of each pair and are not obtained by subtracting two baseline-relative BDBRs.

\begin{table*}[!htbp]
\centering
\caption{Conditioning ablation, BDBR (\%). \emph{Base}: relative to direct
compression; \emph{SCOPE}: pairwise relative to the full model. Lower is
better. Bold/underline indicate the best/second-best Base means.
$^\dagger$: available comparisons only.}
\label{tab:conditioning_all}

\scriptsize
\setlength{\tabcolsep}{2.4pt}
\renewcommand{\arraystretch}{0.88}

\begin{tabular}{
ll
r
@{\hspace{8pt}}
r@{\hspace{0pt}}r
@{\hspace{8pt}}
r@{\hspace{0pt}}r
@{\hspace{8pt}}
r@{\hspace{0pt}}r
@{\hspace{8pt}}
r@{\hspace{0pt}}r
}
\toprule
& &
\multicolumn{1}{c}{SCOPE} &
\multicolumn{2}{c}{Family} &
\multicolumn{2}{c}{Severity} &
\multicolumn{2}{c}{Per-codec} &
\multicolumn{2}{c}{Per-codec+Sev.} \\
\cmidrule(lr){3-3}
\cmidrule(lr){4-5}
\cmidrule(lr){6-7}
\cmidrule(lr){8-9}
\cmidrule(lr){10-11}

Codec & Detector &
Base &
Base & SCOPE &
Base & SCOPE &
Base & SCOPE &
Base & SCOPE \\
\midrule

JPEG & Faster R-CNN
& -15.88
& -14.90 & +0.96
& -2.94 & +14.95
& +4.61 & +22.34
& +4.90 & +22.09 \\

JPEG & RT-DETR-l
& -5.71
& -6.58 & -0.70
& -1.21 & +3.79
& -- & +71.19
& -- & +73.72 \\

JPEG & YOLO26x
& -14.92
& -13.84 & +1.11
& -2.51 & +14.30
& +5.64 & +19.93
& +5.05 & +18.98 \\

\cmidrule(lr){2-11}

JPEG & \textit{Mean}
& \textbf{-12.17}
& \underline{-11.77} & +0.46
& -2.22 & +11.01
& +5.13$^\dagger$ & +37.82
& +4.98$^\dagger$ & +38.26 \\

\midrule

BPG & Faster R-CNN
& -21.74
& -20.27 & +1.85
& -21.07 & +0.99
& -21.54 & +0.20
& -20.42 & +1.86 \\

BPG & RT-DETR-l
& -19.51
& -17.87 & +2.04
& -17.69 & +2.37
& -19.98 & -0.60
& -18.35 & +1.41 \\

BPG & YOLO26x
& -20.88
& -20.40 & +0.47
& -20.01 & +0.95
& -21.22 & -0.60
& -18.91 & +2.42 \\

\cmidrule(lr){2-11}

BPG & \textit{Mean}
& \underline{-20.71}
& -19.51 & +1.45
& -19.59 & +1.44
& \textbf{-20.91} & -0.33
& -19.23 & +1.90 \\

\midrule

BMSHJ & Faster R-CNN
& -19.21
& -17.28 & +1.79
& -21.25 & -2.80
& -20.40 & -2.16
& -24.65 & -6.46 \\

BMSHJ & RT-DETR-l
& -13.08
& -12.58 & +0.88
& -16.19 & -3.59
& -16.27 & -3.99
& -16.64 & -3.97 \\

BMSHJ & YOLO26x
& -12.91
& -15.09 & -3.18
& -16.62 & -4.56
& -17.50 & -6.01
& -18.36 & -5.92 \\

\cmidrule(lr){2-11}

BMSHJ & \textit{Mean}
& -15.07
& -14.98 & -0.17
& -18.02 & -3.65
& \underline{-18.06} & -4.05
& \textbf{-19.88} & -5.45 \\

\midrule

\multicolumn{2}{l}{Overall}
& \textbf{-15.98}
& \underline{-15.42} & +0.58
& -13.28 & +2.93
& -13.33$^\dagger$ & +11.14
& -13.42$^\dagger$ & +11.57 \\

\bottomrule
\end{tabular}
\end{table*}

Full SCOPE gives the best overall result at $-15.98\%$ BDBR. Family-only is closest at $-15.42\%$, whereas severity-only reaches $-13.28\%$, indicating that codec family is the stronger global conditioning signal. Adding severity nevertheless yields the best aggregate result, although its effect is not uniform across codecs. Codec-specific specialists can be stronger locally. Per-codec+Sev.~reaches $-19.88\%$ on BMSHJ-but are substantially less consistent and reach $+4.98\%$ on JPEG over the available comparisons. The shared family-and-severity model therefore provides the strongest overall trade-off while retaining one set of weights.
\section{Conclusion}
\label{sec:conclusion}

We introduced SCOPE, a conditional image pre-editor for object detection under image compression. The model combines codec-family and severity conditioning with attention-based object/background editing, allowing one set of weights to adapt its transformation across multiple compression regimes. Training uses differentiable codec proxies and a frozen detector, while inference requires only the image and codec condition and leaves both the actual codec and downstream detector unchanged.

Across JPEG, BPG, and BMSHJ and the three evaluated detector architectures, SCOPE achieves an average BDBR of $-15.98\%$ relative to direct compression. The conditioning ablation shows that family information provides the strongest individual conditioning signal, while combining family and severity gives the best aggregate result with a single model. SCOPE also transfers zero-shot to the unseen Cheng and JPEG2000 codecs, although the JPEG2000 results show that the chosen seen-family condition can affect performance on an unseen family.

These results support conditional pre-editing as a practical, standard-compatible strategy for machine-oriented compression: adaptation is moved in front of the codec, preserving conventional compressed images while improving the rate-accuracy trade-off for downstream detection.

%
%
\bibliographystyle{splncs04}
\bibliography{main}

\clearpage
\appendix
\section*{Supplementary Material}















\section{Runtime Analysis}
\label{suppl:runtime_analysis}

We measure the runtime overhead introduced by SCOPE independently from codec execution and detector evaluation. The benchmark is performed on the COCO 2017~\cite{lin2014microsoft} Validation set using the same checkpoint used in the main experiments. For each codec-family condition, namely \texttt{Diff\_JPEG}, \texttt{BPG\_like}, and \texttt{BMSHJ}, we evaluate five severity values $s \in \{0.00, 0.25, 0.50, 0.75, 1.00\}$. Each family-severity pair is measured on 5,000 validation images, resulting in 75,000 total pre-editing measurements. We use 20 warm-up images before timing and one timed pass per image.

The reported runtime isolates the pre-editing module. Image reading is measured separately and excluded from the reported pre-editing time, while codec execution and detector evaluation are not performed. Output saving is disabled during the benchmark, so the total pre-editing time includes tensor preparation, transfer to the GPU, construction of the conditioning inputs, and the forward pass of the pre-editor. The pre-editor is evaluated on CUDA with batch size one.

Table~\ref{tab:runtime_by_family} summarizes the runtime averaged on all severity levels for each codec-family condition. The forward pass takes about $7.93$ ms per image on average, corresponding to approximately $126$ images per second. Including tensor preparation and conditioning setup, the total pre-editing time is about $11.81$ ms per image, corresponding to approximately $85$ images per second. The differences across codec-family conditions are very small, which is expected because SCOPE uses the same network architecture for all conditions; the codec family and severity only modify the conditioning vector.

\begin{table}[!htbp]
\centering
\scriptsize
\caption{
Runtime of SCOPE grouped by codec-family condition. Values are averaged over five severity levels and validation images per level. Forward time includes only the neural pre-editor forward pass, while total time additionally includes tensor preparation and conditioning setup. Codec execution, detector evaluation, and image reading are excluded.
}
\label{tab:runtime_by_family}
\begin{tabular}{lrrrrrr}
\toprule
Condition family & Images & Forward mean & Forward p95 & Total mean & Total p95 & Total FPS \\
\midrule
\texttt{BMSHJ}      & 25,000 & 7.92 ms & 10.69 ms & 11.78 ms & 15.13 ms & 84.89 \\
\texttt{BPG\_like}  & 25,000 & 7.92 ms & 10.67 ms & 11.79 ms & 15.11 ms & 84.85 \\
\texttt{Diff\_JPEG} & 25,000 & 7.96 ms & 10.85 ms & 11.85 ms & 15.17 ms & 84.37 \\
\midrule
Overall             & 75,000 & 7.93 ms & 10.74 ms & 11.81 ms & 15.14 ms & 84.70 \\
\bottomrule
\end{tabular}
\end{table}

We also analyze whether the conditioning severity affects runtime. As shown in Table~\ref{tab:runtime_by_severity}, the total runtime remains nearly constant across severity values. The small variation is within normal measurement noise and does not indicate a systematic dependence on the severity parameter. This confirms that changing the compression severity only changes the scalar conditioning input and does not alter the computational graph of the pre-editor.

\begin{table}[!htbp]
\centering
\scriptsize
\caption{
Runtime of SCOPE grouped by severity. Values are averaged over the three codec-family conditions. Runtime is nearly constant across severity values because severity only affects the conditioning vector.
}
\label{tab:runtime_by_severity}
\begin{tabular}{lrrrr}
\toprule
Severity & Images & Forward mean & Total mean & Total FPS \\
\midrule
0.00 & 15,000 & 8.00 ms & 11.87 ms & 84.28 \\
0.25 & 15,000 & 7.92 ms & 11.78 ms & 84.89 \\
0.50 & 15,000 & 7.91 ms & 11.79 ms & 84.78 \\
0.75 & 15,000 & 7.92 ms & 11.79 ms & 84.82 \\
1.00 & 15,000 & 7.92 ms & 11.80 ms & 84.73 \\
\bottomrule
\end{tabular}
\end{table}

Overall, the runtime analysis shows that SCOPE introduces a small and stable pre-processing overhead. The cost is effectively independent of the selected codec-family condition and severity level because all operating points are handled by the same conditional network. This supports the practical use of SCOPE as a codec-conditioned pre-editing module: the same model can be applied across different compression settings by changing only the family and severity conditions, without modifying the architecture, loading codec-specific weights, or increasing the inference cost.

\section{Semantic Decomposition of mAP Gains}
\label{suppl:semantic_decomposition}

We further analyze which semantic groups contribute to the detection gains obtained by SCOPE. 
The goal of this analysis is to decompose the same COCO mAP difference used in the main rate and accuracy curves, rather than to report independent AP values for each supercategory.

For this reason, we do not compute the difference between AP values evaluated directly on each supercategory. 
Instead, we first compute AP for each individual COCO category and then aggregate the category-level differences according to the official COCO supercategories. 
Let $\mathcal{C}$ be the complete set of COCO categories and let $\mathcal{C}_s \subset \mathcal{C}$ be the set of categories belonging to the supercategory $s$. 
For a fixed codec setting and detector, the overall mAP difference is

\begin{equation}
    \Delta \mathrm{mAP}
    =
    \frac{1}{|\mathcal{C}|}
    \sum_{c \in \mathcal{C}}
    \left(
        \mathrm{AP}^{\mathrm{SCOPE}}_c
        -
        \mathrm{AP}^{\mathrm{base}}_c
    \right).
\end{equation}

The contribution of the supercategory $s$ is then defined as

\begin{equation}
    \Delta \mathrm{mAP}_s
    =
    \frac{1}{|\mathcal{C}|}
    \sum_{c \in \mathcal{C}_s}
    \left(
        \mathrm{AP}^{\mathrm{SCOPE}}_c
        -
        \mathrm{AP}^{\mathrm{base}}_c
    \right).
\end{equation}

The denominator is therefore the total number of COCO categories, not the number of categories inside the supercategory. 
Consequently, the contributions recover the overall mAP difference:

\begin{equation}
    \Delta \mathrm{mAP}
    =
    \sum_s \Delta \mathrm{mAP}_s .
\end{equation}

This decomposition should therefore be interpreted as a breakdown of the global COCO mAP variation. 
A positive bar indicates that the corresponding supercategory contributes positively to the overall gain, while a negative bar indicates that SCOPE decreases the category-averaged AP for that semantic group at that operating point. 
Individual negative bars are possible even when the total mAP improves, as long as the sum of all contributions remains positive.

Table~\ref{tab:coco_supercategory_groups} reports the category grouping used in the decomposition. 
We use the official COCO supercategories. 
For single-category groups, we report only the group name.

\begin{table}[!htbp]
\centering
\scriptsize
\caption{
COCO category grouping used for the semantic decomposition.
}
\label{tab:coco_supercategory_groups}
\begin{tabularx}{\textwidth}{p{0.18\textwidth}X}
\toprule
Group & COCO categories \\
\midrule
person & person \\
vehicle & bicycle, car, motorcycle, airplane, bus, train, truck, boat \\
outdoor & traffic light, fire hydrant, stop sign, parking meter, bench \\
animal & bird, cat, dog, horse, sheep, cow, elephant, bear, zebra, giraffe \\
accessory & backpack, umbrella, handbag, tie, suitcase \\
sports & frisbee, skis, snowboard, sports ball, kite, baseball bat, baseball glove, skateboard, surfboard, tennis racket \\
kitchen & bottle, wine glass, cup, fork, knife, spoon, bowl \\
food & banana, apple, sandwich, orange, broccoli, carrot, hot dog, pizza, donut, cake \\
furniture & chair, couch, potted plant, bed, dining table, toilet \\
electronic & tv, laptop, mouse, remote, keyboard, cell phone \\
appliance & microwave, oven, toaster, sink, refrigerator \\
indoor & book, clock, vase, scissors, teddy bear, hair drier, toothbrush \\
\bottomrule
\end{tabularx}
\end{table}

We report this analysis only for BMSHJ and BPG. 
These codecs provide more reliable paired comparisons between the compressed baseline and the SCOPE-preedited version at the same nominal compression setting. 
For JPEG, the resulting bitrates of the baseline and SCOPE variants can differ substantially more at the same quality value. 
A same-quality semantic decomposition would therefore not align as cleanly with the rate and accuracy curves, since the comparison would mix the semantic effect of pre-editing with a larger bitrate shift. 
For this reason, JPEG is kept in the global rate and accuracy analysis but excluded from the semantic decomposition plots.

Figures~\ref{fig:suppl_decomp_bmshj} and~\ref{fig:suppl_decomp_bpg} show the semantic decomposition of BMSHJ and BPG. 
Overall, the histograms indicate that the gains are not concentrated in a single semantic group. 
Under stronger compression, most detectors exhibit positive contributions across several groups, especially food, animal, electronics, furniture, kitchen, and indoor categories. 
This suggests that SCOPE helps preserve task-relevant structures that are useful for detection across heterogeneous object types.

The effect becomes smaller as compression becomes lighter. 
This is expected because the compressed baseline already preserves most of the information needed by the detector at low compression. 
In this regime, the histograms often show near-zero contributions and a mixture of positive and negative bars. 
For BMSHJ at low compression, the total effect is close to neutral, with a small positive or nearly unchanged result for most detectors and a slight decrease for RT-DETR. 
For BPG at low compression, the same-quality decomposition is slightly negative for all detectors. 
Therefore, these low-compression results should be interpreted as comparable to the baseline rather than as a failure mode of the method.

This interpretation is also consistent with the main paper results in Fig.~3. 
At low compression, the main advantage of SCOPE is not necessarily a large same-quality mAP increase, but the ability to obtain comparable detection accuracy at a lower bitrate. 
Therefore, a small negative or near-zero same-quality contribution in the decomposition does not contradict the rate and accuracy curves. 
It indicates that, when the baseline is already strong, SCOPE mainly shifts the operating point toward lower bits per pixel while keeping the detector performance close to the baseline.

For medium and high compression, the decompositions are more clearly positive. 
BMSHJ shows the largest and most consistent gains at high compression, where compression artifacts are strongest and the pre-editor has more room to improve detector-relevant content. 
BPG shows positive gains at medium and high compression, while becoming nearly neutral or slightly negative at low compression. 
Across both codecs, the recurrent positive contributions from object-rich groups such as food, animal, electronics, furniture, kitchen, and indoor suggest that SCOPE is especially helpful when compression affects localized structures and textures that are important for object recognition. 
Conversely, groups such as sports, outdoor, and vehicle are more variable across detectors and compression levels, indicating that the effect of pre-editing can depend on the detector architecture and on the visual statistics of each semantic group.

\setcounter{figure}{6}

\begin{figure*}[t]
    \centering

    \begin{subfigure}[t]{0.95\textwidth}
        \centering
        \includegraphics[width=\linewidth]{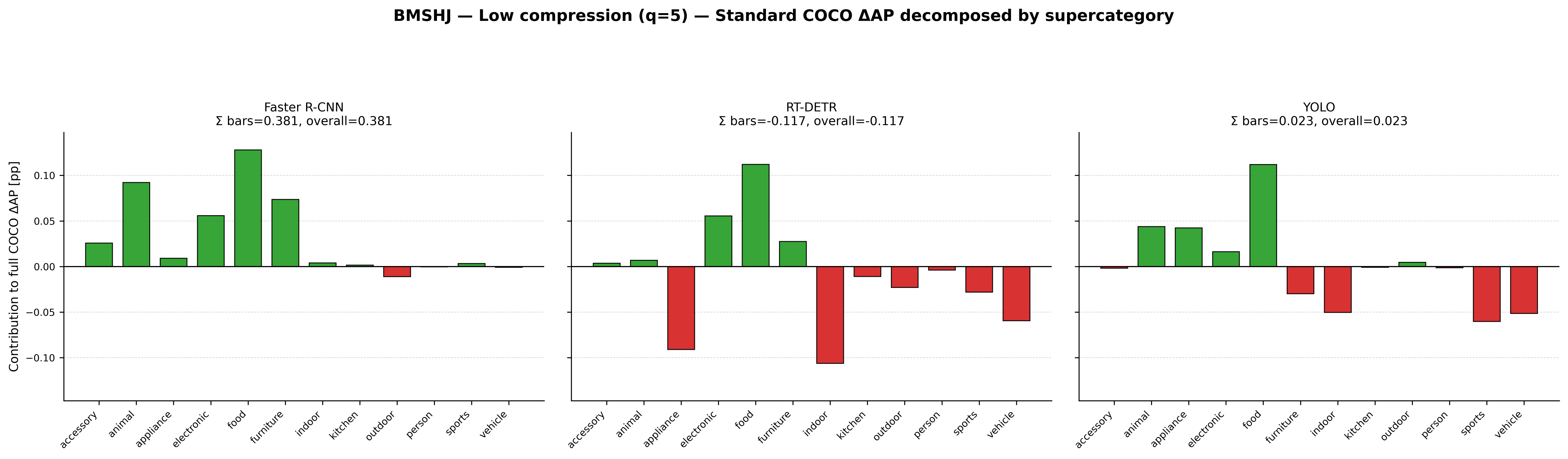}
        \caption{BMSHJ, low compression.}
        \label{fig:suppl_decomp_bmshj_low}
    \end{subfigure}

    \vspace{0.4em}

    \begin{subfigure}[t]{0.95\textwidth}
        \centering
        \includegraphics[width=\linewidth]{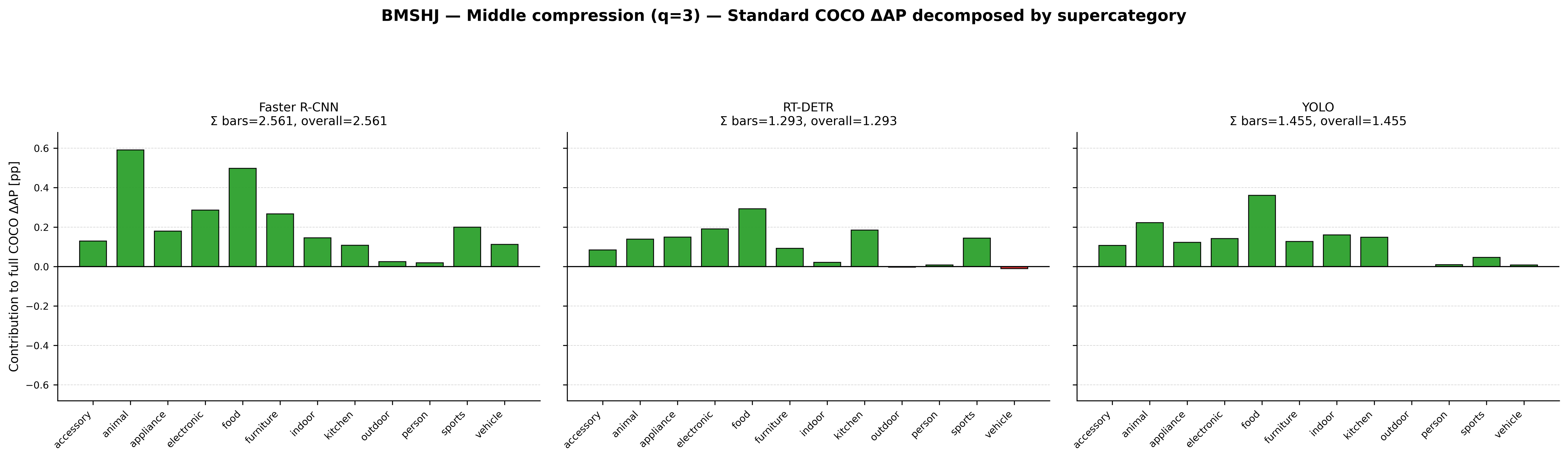}
        \caption{BMSHJ, middle compression.}
        \label{fig:suppl_decomp_bmshj_mid}
    \end{subfigure}

    \vspace{0.4em}

    \begin{subfigure}[t]{0.95\textwidth}
        \centering
        \includegraphics[width=\linewidth]{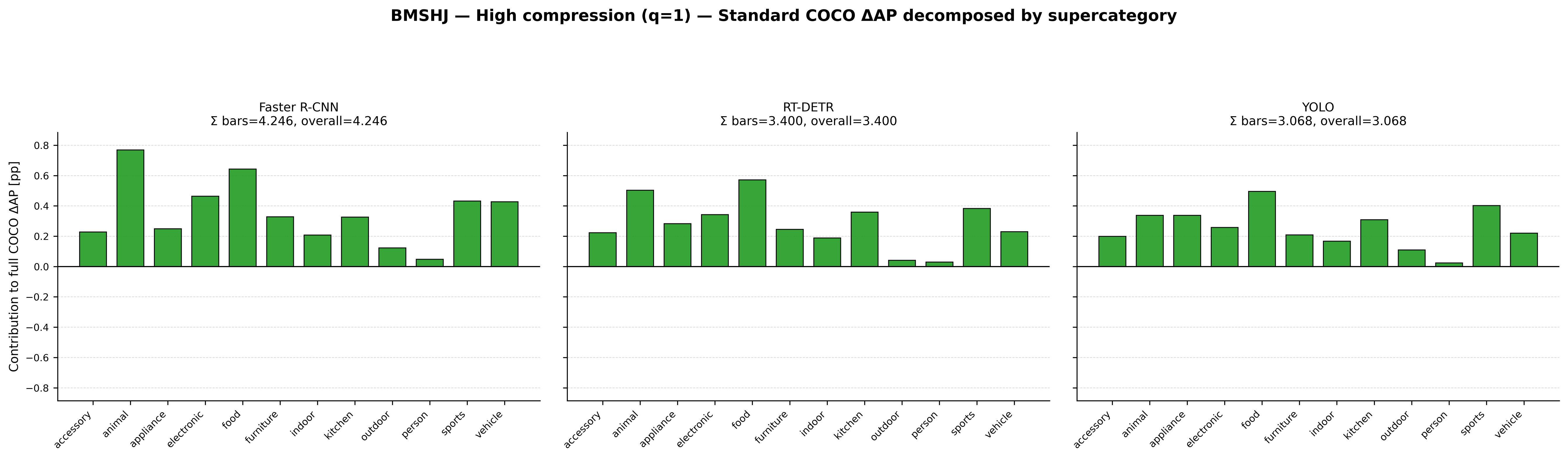}
        \caption{BMSHJ, high compression.}
        \label{fig:suppl_decomp_bmshj_high}
    \end{subfigure}

    \caption{
    Semantic decomposition of the COCO mAP difference for BMSHJ.
    Each bar represents the contribution of a COCO supercategory to the overall mAP change between SCOPE and the corresponding compressed baseline.
    Contributions are obtained from per-category AP differences and normalized by the total number of COCO categories, so the sum of the bars equals the overall mAP difference for the corresponding detector and compression setting.
    }
    \label{fig:suppl_decomp_bmshj}
\end{figure*}

\begin{figure*}[t]
    \centering

    \begin{subfigure}[t]{0.95\textwidth}
        \centering
        \includegraphics[width=\linewidth]{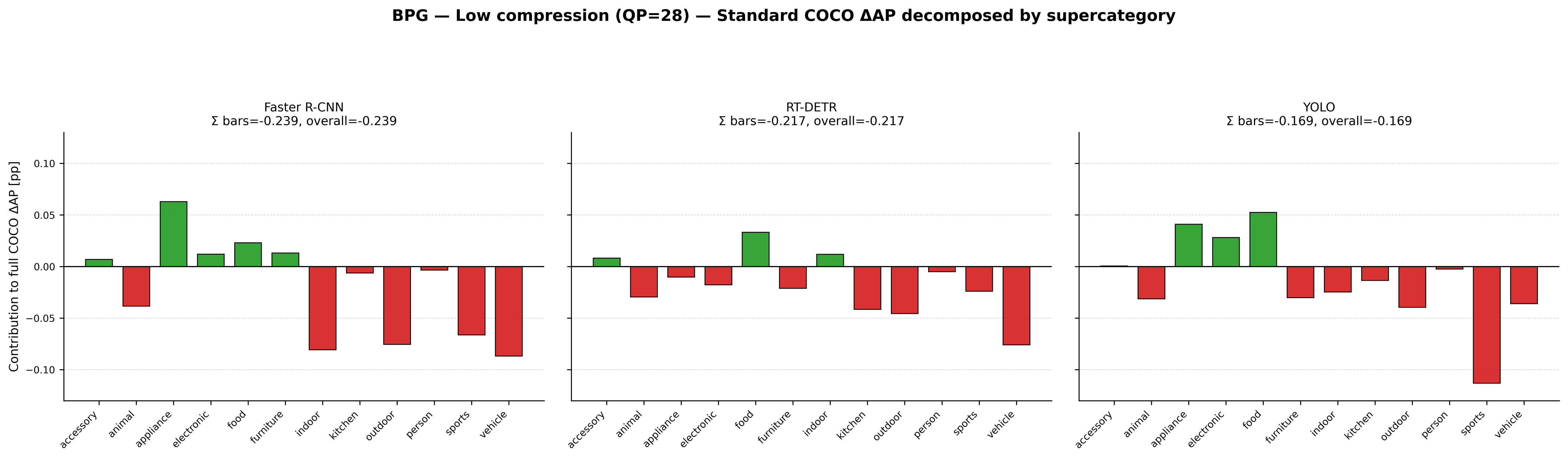}
        \caption{BPG, low compression.}
        \label{fig:suppl_decomp_bpg_low}
    \end{subfigure}

    \vspace{0.4em}

    \begin{subfigure}[t]{0.95\textwidth}
        \centering
        \includegraphics[width=\linewidth]{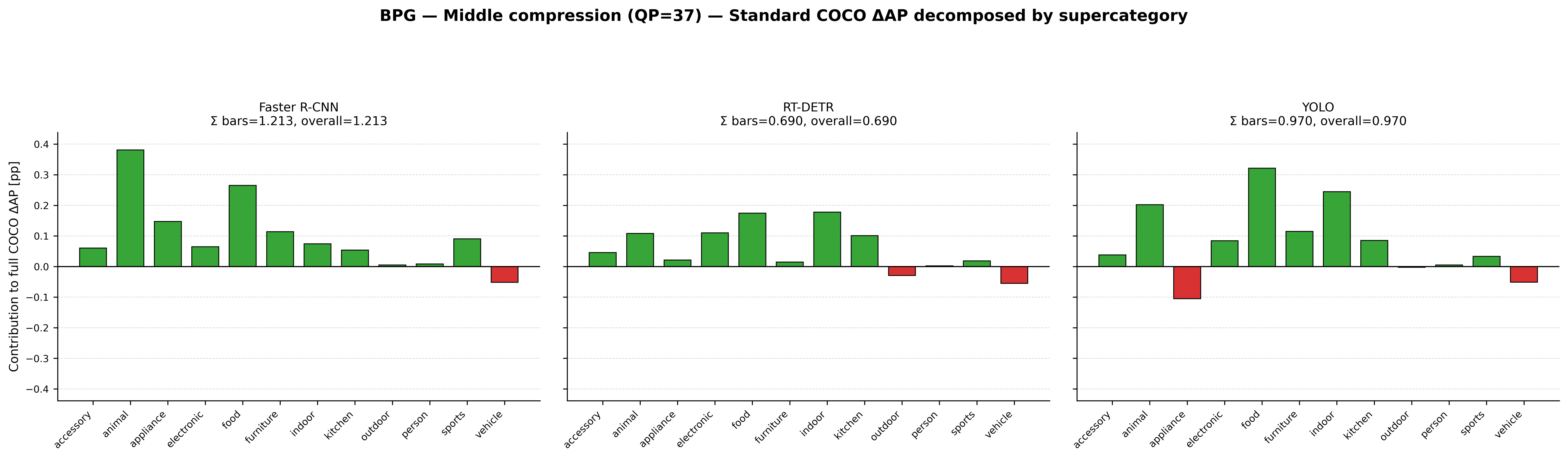}
        \caption{BPG, middle compression.}
        \label{fig:suppl_decomp_bpg_mid}
    \end{subfigure}

    \vspace{0.4em}

    \begin{subfigure}[t]{0.95\textwidth}
        \centering
        \includegraphics[width=\linewidth]{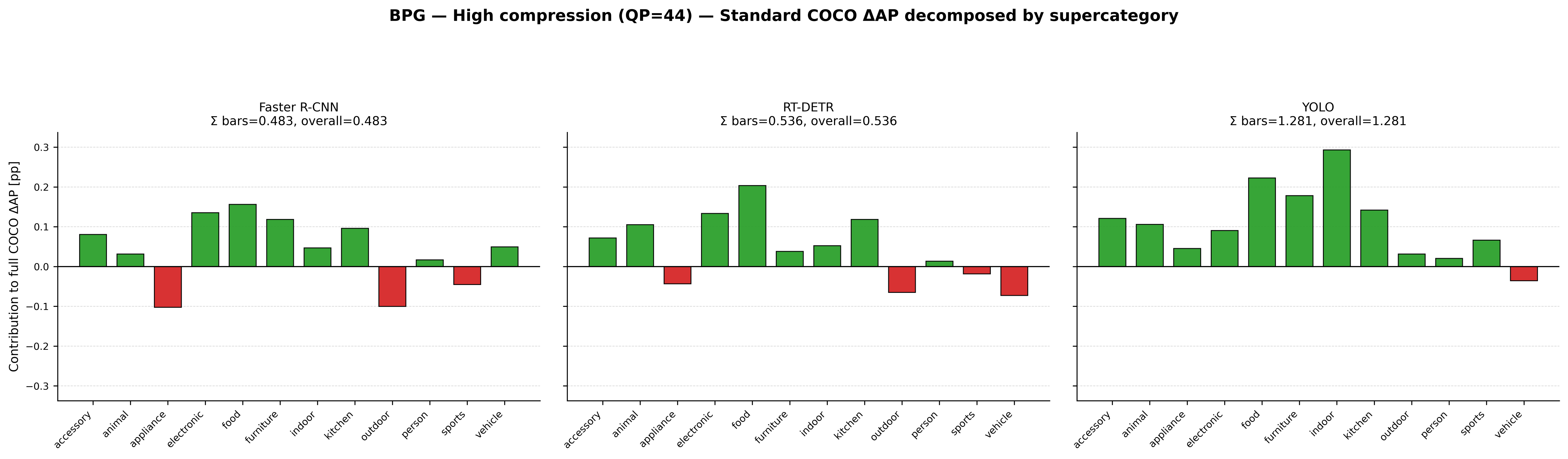}
        \caption{BPG, high compression.}
        \label{fig:suppl_decomp_bpg_high}
    \end{subfigure}

    \caption{
    Semantic decomposition of the COCO mAP difference for BPG.
    As in Fig.~\ref{fig:suppl_decomp_bmshj}, the decomposition is computed from per-category AP differences and then grouped by COCO supercategory using the full COCO category set as denominator.
    This makes the sum of the displayed contributions directly comparable to the overall mAP difference shown in the rate and accuracy curves.
    }
    \label{fig:suppl_decomp_bpg}
\end{figure*}




\end{document}